\documentclass[lettersize,journal]{IEEEtran}
\usepackage{amsmath,amsfonts}
\usepackage{algorithmic}
\usepackage{algorithm}
\usepackage{array}
\usepackage[caption=false,font=normalsize,labelfont=sf,textfont=sf]{subfig}
\usepackage{textcomp}
\usepackage{stfloats}
\usepackage{url}
\usepackage{verbatim}
\usepackage{graphicx}
\usepackage{cite}
\usepackage{multirow}
\usepackage{graphicx}
\usepackage{booktabs}
\usepackage{tabularx}
\usepackage{mathtools}
\usepackage{xcolor}
\usepackage{tikz}
\begin{document}

\title{MambaLoc: Efficient Camera Localisation \\ via State Space Model}


\author{Jialu Wang$^{1}$, Kaichen Zhou$^{1}$, Andrew Markham $^{1}$ and Niki Trigoni $^{1}$
\thanks{$^{1}$Jialu Wang, Kaichen Zhou, Niki Trigoni and Andrew Markham are with Department of Computer Science,
	University of Oxford,
	UK, 
    {\tt\small jialu.wang@cs.ox.ac.uk},
     {\tt\small kc.zhou2020@hotmail.com}, 
    	{\tt\small niki.trigoni@cs.ox.\newline ac.uk}, 
    		{\tt\small andrew.markham@cs.ox.ac.uk}},
}




\maketitle

\begin{abstract}

Location information is pivotal for the automation and intelligence of terminal devices and edge-cloud IoT systems, such as autonomous vehicles and augmented reality, while achieving reliable positioning across diverse IoT applications remains challenging due to significant training costs and the necessity of densely collected data. 
To tackle these issues, we have innovatively applied the selective state space (SSM) model to visual localization, introducing a new model named MambaLoc. 
The proposed model demonstrates exceptional training efficiency by capitalizing on the SSM model's strengths in efficient feature extraction, rapid computation, and memory optimization, and it further ensures robustness in sparse data environments due to its parameter sparsity. 
Additionally, we propose the Global Information Selector (GIS), which leverages selective SSM to implicitly achieve the efficient global feature extraction capabilities of Non-local Neural Networks. 
This design leverages the computational efficiency of the SSM model alongside the Non-local Neural Networks’ capacity to capture long-range dependencies with minimal layers. Consequently, the GIS enables effective global information capture while significantly accelerating convergence. 
Our extensive experimental validation using public indoor and outdoor datasets first demonstrates our model's effectiveness, followed by evidence of our GIS's versatility with various existing localization models. For instance, in our experiment with the Heads Scene of the 7Scenes dataset \cite{glocker2013real}, MambaLoc used just 0.05\% of the training set and achieved accuracy comparable to the leading methods in just 22.8 seconds. We also provide a demo of MambaLoc deployed on end-user devices. These results highlight MambaLoc's potential as a robust and efficient solution for visual localization in edge-cloud IoT and terminal devices, significantly enhancing the commercial viability of deep neural networks for camera localization. Our code and models are publicly available to support further research and development in this area. 

\end{abstract}


\section{Introduction}

\begin{figure*}[h]
    \centering
    \caption{\textbf{Proposed Bidirectional Global Information Selector (GIS) (right)}: The GIS starts by concatenating the visual encoder's output, \( G_{in}\), with its flipped version, \( G_{flip} \). This combined input, \( G_{concat} \), is then processed by a bidirectional single-layer Mamba Module, which selectively compresses it using a hardware-aware algorithm. \textbf{Architecture of MambaLoc (left)}: MambaLoc consists of a shared CNN backbone and two distinct branches designed to independently regress the camera's position and orientation. Each branch includes a dedicated Transformer-Encoder, a  Global Information Selector (GIS), and an MLP head.
}
    \includegraphics[width=\textwidth]{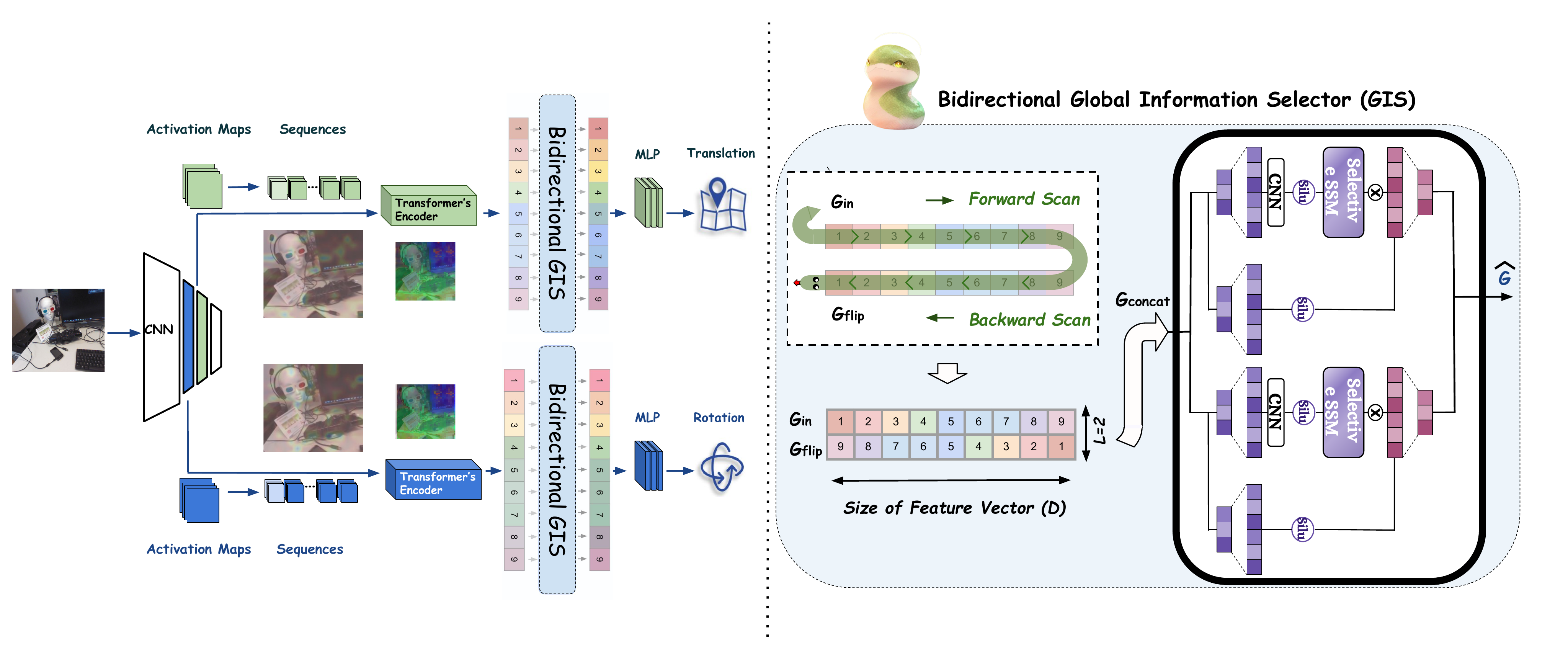}

    \label{fig:MambaNonLocal}
\end{figure*}

Deep learning-based camera localization employs neural networks to implicitly learn a map of a given scene, allowing for the estimation of 6-DoF of camera pose (i.e., 6 degrees of freedom, including three translations and three rotations) for a moving camera based on captured images. It is fundamental and crucial for the automation and intelligent systems in Internet of Things (IoT), such as autonomous vehicles, emergency response devices, augmented reality, domestic robot navigation, and intelligent decision-making robots \cite{xu2019ivr}, etc. This problem was initially solved as a place recognition problem~\cite{zhang2006image, robertson2004image, philbin2007object, shen2013spatially, chum2011total, tolias2014visual} which estimate the pose of the query image by retrieving the most similar images from the database. However, the robustness of these retrieval-based \cite{irschara2007towards} methods are limited by its hand-crafted local features \cite{irschara2009structure}. To further improve the estimation accuracy, structure based methods proposed to recover the camera pose by finding correspondence between query images and the 3D scene model. Although these methods achieves state-of-art (SOTA) accuracy~\cite{xu2022critical}, they are resource-intensive and requires pre-obtained SfM models or highly accurate depth information, which is not universally available in lightweight customer devices such as laptops and smartphones.  

Recent developments in deep neural networks (DNN) for image processing have facilitated the emergence of absolute pose regression (APR) methods, capable of directly predicting camera poses from images via a singular DNN \cite{kendall2015posenet,wang2020atloc,zhou2021vmloc}. Despite being less accurate than structure-based methods, APR-based methods are lightweight, fast, and require no additional data during inference. To better explore the application of localization methods in terminal devices and edge IoT systems, this study focuses on such localization methods. Nonetheless, these methods encounter two principal obstacles: \textbf{i) Trade-off Between Accuracy and Efficiency}: Most model development has focused on CNN-based or Transformer-based designs. CNNs are effective at extracting local features but often sacrifice resolution, while Transformers offer a global perspective but increase computational demands. Existing methods typically require hours \cite{shavit2024learning} or even a day \cite{chen2021direct, chen2022dfnet} to train a single scene. This trade-off between accuracy and efficiency remains a significant challenge; \textbf{ii) Lack of Robustness to Sparse Viewpoints}: These data-driven methods heavily rely on dense training samples, limiting their practical applications, such as requiring low data exchange costs for cloud-edge integrated IoT localization or using sparse keyframes for loop detection in SLAM on end-user devices. 

Recently, state space models such as Mamba \cite{gu2023mamba} have demonstrated substantial global modeling capabilities across diverse domains, including language modeling and computer vision \cite{zhu2024vision}. These models are particularly notable for reducing the time complexity of global information extraction to $O(\mathbf{N})$, marking a significant stride in computational efficiency. Inspired by this, our study explores the potential of state space models in visual localization and introduces a highly efficient model named MambaLoc. MambaLoc leverages Mamba's strengths to achieve efficient feature extraction, rapid computation, and optimized memory usage, while also enhancing robustness, particularly in environments with sparse training samples. Specifically, we introduce the Global Information Selector (GIS), which combines the global information aggregation capabilities of non-local mechanisms \cite{wang2018non} with Mamba's expertise, further enhancing the model's overall performance. We leverage Mamba's learnable parameters to selectively compress input features into a more effective global feature representation.
This combination enables a single Mamba layer to function as a Non-local Neural Network, effectively capturing long-range dependencies with minimal layers and accelerating convergence. As a versatile block, GIS can be seamlessly added to existing localization models with just a few lines of code, greatly reducing training time and lessening the dependency on densely collected training data. Through extensive experiments on a variety of indoor and outdoor public datasets, we initially demonstrated the effectiveness of MambaLoc, subsequently showcasing the adaptability of our GIS across multiple existing localization models. For instance, on the 7Scenes Dataset \cite{glocker2013real}, MambaLoc achieved state-of-the-art translation accuracy in only 22.8 seconds using merely 0.05\% of the training samples, matching the pose estimation accuracy of leading models. Additionally, we provide experimental evidence supporting the practical deployment of our model on terminal devices. These findings highlight MambaLoc's potential as a robust and efficient solution for visual localization within neural networks, enhancing location services for terminal devices and edge-cloud integrated IoT. In conclusion, our contributions are as follows:
\begin{enumerate}
    \item We have innovatively integrated the Mamba framework into visual localization, creating MambaLoc, a new model that demonstrates exceptional training efficiency and robustness in sparse data environments.
    
    \item We propose the versatile Global Information Selector (GIS), which leverages selective SSM to implicitly achieve the efficient global feature extraction capabilities of Non-local Neural Networks. This Block can be seamlessly integrated into existing localization models with minimal code changes, significantly reducing training time and dependency on densely collected data.

    \item We provide extensive experimental validation using public indoor and outdoor datasets, demonstrating the effectiveness of our MambaLoc model and the versatility of our GIS with various existing localization models. Additionally, we showcase a demo of MambaLoc deployed on end-user devices. 

\end{enumerate}



\section{Related Work}
\subsection{Deep Neural Networks For Camera localization}

Visual localization involves constructing a scene representation from a set of mapping images and their corresponding poses within a common coordinate system. Given a query image, the objective is to estimate its pose (position and orientation) relative to the scene. Existing solutions to this problem can be broadly categorized into two primary approaches: structure-based techniques and Absolute Pose Regression (APR) techniques.

\textbf{Structure-Based Techniques} rely on geometric methods to establish correspondences between images and the scene map. These methods achieve state-of-the-art accuracy \cite{xu2022critical}, but they are resource-intensive, requiring pre-computed Structure-from-Motion (SfM) models or highly accurate depth information. Such dependencies make these methods less practical for lightweight consumer devices, such as laptops and smartphones.

\textbf{Absolute Pose Regression (APR) techniques} take a different approach by leveraging neural networks to directly predict the absolute position and orientation of an image, bypassing the need for explicit matching between the 2D image and the 3D scene map. The pioneering work in this domain, PoseNet, introduced by Kendall et al. \cite{kendall2015posenet}, employs a feed-forward neural network to predict a 7-dimensional pose vector for each input image. Subsequent research has introduced several architectural innovations, including hourglass networks \cite{melekhov2017image}, bifurcated translation and rotation regression \cite{wu2017delving,naseer2017deep}, attention mechanisms \cite{zhou2020kfnet, shavit2024learning}, and LSTM layers \cite{walch2017image}. Efforts to further enhance APR performance have explored diverse supervision techniques, such as geometric loss \cite{brahmbhatt2018geometry,zhou2023mgdepth}, relative pose constraints \cite{wang2024wscloc}, uncertainty modeling \cite{vaghi2022uncertainty}, and sequential formulations like temporal filtering \cite{zhou2020kfnet} and multitasking \cite{shavit2024learning}. While APR methods generally offer lower accuracy compared to structure-based approaches, they are advantageous for being lightweight, fast, and not requiring additional data during inference.

To address overfitting in APR methods, recent research has explored novel view synthesis techniques that generate large volumes of training data \cite{purkait2018synthetic, moreau2022lens, chen2022dfnet, zhao2024pnerfloc, zhou2024dynpoint}. However, generating high-quality synthetic data is both time-consuming and resource-intensive, with significant memory and data exchange costs, rendering these methods less suitable for deployment on edge or terminal devices.

Given the increasing relevance of edge computing and the constraints of terminal devices, this study focuses on optimizing APR methods to make them more efficient and practical for use in such environments.


\subsection{ State Space Models for Sequence Modeling}
State space models have been proposed as effective tools for sequence modeling. The Structured State-Space Sequence (S4) model \cite{gu2021efficiently} offers a novel alternative to CNNs and Transformers for capturing long-range dependencies, with the added benefit of linear scalability in sequence length, prompting further research. Building on this, \cite{smith2022simplified} introduces the new S5 layer, incorporating MIMO SSM and efficient parallel scanning into the S4 layer. Similarly, \cite{fu2022hungry} develops the H3 SSM layer, significantly closing the performance gap between SSMs and Transformer attention in language modeling. \cite{mehta2022long} enhances the S4 model with the Gated State Space layer by adding more gating units to improve expressivity. Recently, \cite{gu2023mamba} presents a data-dependent SSM layer and constructs Mamba, a generic language model backbone that outperforms Transformers at various sizes on large-scale real-world data while maintaining linear scaling in sequence length. Inspired by Mamba's success, our work focuses on applying these advancements to camera localization, aiming to create an efficient and high-performance generic camera localization backbone.


\section{Methodology}

The goal of MambaLoc is to utilizes an advanced selective state space model (SSM), specifically Mamba, to implicitly replicate the efficient global feature extraction capabilities of Non-local Neural Networks. This integration allows the model to efficiently capture global information while significantly improving training efficiency. Ultimately, MambaLoc can be deployed on Edge-Cloud Collaborative IoT and Terminal Devices, greatly enhancing the commercial viability of deep neural networks for camera localization.

This section begins with a description of the preliminaries of Sequence Modeling for Camera Localization, particularly focusing on Mamba. It proceeds with a comprehensive overview of MambaLoc, outlining its architectural design. Subsequently, we elucidate the integration of Non-local Neural Network concepts with Mamba in our Global Information Selector (GIS). Finally, we discuss the challenges encountered when decoupling the localization model from cloud-based systems and detail our deployment strategies for MambaLoc on edge and terminal devices.

\subsection{Preliminaries: Sequence Modeling for Camera localization}

In this paragraph, we review existing deep learning-based camera localization models from the perspective of sequence modeling. While RNNs possess the capability to handle sequences, they encounter difficulties with gradient vanishing when processing long sequences. LSTM, on the other hand, excels in capturing long-term dependencies in data \cite{walch2017image}. CNNs, compared to LSTM, are more effective at capturing local features and offer stronger parallel computing capabilities \cite{kendall2015posenet}. Transformers, known for their robust parallel computation and ability to handle long-range dependencies across various sequence tasks, face challenges with long sequences due to their quadratic time complexity relative to sequence length \cite{wang2020atloc}.

Recently, Albert Gu et al. introduced Mamba \cite{gu2023mamba}, a novel approach leveraging a selective state-space model architecture to overcome the limitations of traditional methods, thereby enhancing performance, efficiency, and scalability in academic contexts. Inspired by continuous systems, structured state space sequence models (SSMs) such as S4 \cite{gu2021efficiently} and Mamba map a one-dimensional function or sequence \( x(t) \in \mathbb{R} \) to \( y(t) \in \mathbb{R} \) through a hidden state \( h(t) \in \mathbb{R}^N \) (see Equations \ref{eq:11} and \ref{eq:12}). In this framework, \(  \mathbf{A} \in \mathbb{R}^{N \times N} \) serves as the evolution parameter, while \( \mathbf{B} \in \mathbb{R}^{N \times 1} \) and \( \mathbf{C} \in \mathbb{R}^{1 \times N} \) function as the projection parameters.

\begin{equation}
\label{eq:11}
h'(t) = \mathbf{A}h(t) + \mathbf{B}x(t),
\end{equation}

\begin{equation}
\label{eq:12}
y(t) = \mathbf{C}h(t).
\end{equation}

The S4 and Mamba models act as discrete versions of continuous systems by employing a timescale parameter \(\mathbf{\Delta}\) to convert the continuous parameters \(\mathbf{A}\) and \(\mathbf{B}\) into their discrete equivalents, \(\mathbf{\bar{A}}\) and \(\mathbf{\bar{B}}\). A commonly used transformation method is the zero-order hold (ZOH), defined by the following expressions:

\begin{equation}
\label{eq:13}
\mathbf{\bar{A}} = \exp(\mathbf{\Delta A}),
\end{equation}

\begin{equation}
\label{eq:14}
\mathbf{\bar{B}} = (\mathbf{\Delta A})^{-1}(\exp(\mathbf{\Delta A}) - \mathbf{I}) \cdot \mathbf{\Delta B}.
\end{equation}

Once the parameters \(\mathbf{A}\) and \(\mathbf{B}\) are discretized, the discrete form of Equation (1) with a step size \(\mathbf{\Delta}\) can be expressed as:

\begin{equation}
h_t = \mathbf{\bar{A}} h_{t-1} + \mathbf{\bar{B}} x_t,
\end{equation}

\begin{equation}
y_t = \mathbf{C} h_t.
\end{equation}

These equations provide the discrete state \(h_t\) and the corresponding output \(y_t\), effectively bridging the continuous and discrete domains.

\subsection{Architecture of MambaLoc}

MambaLoc consists of a shared convolutional backbone and two branches for regressing the camera's position and orientation separately. Each branch includes an independent Transformer-Encoder, a Global Information Selector (GIS), and an MLP head. By using separate GIS in each branch, we can extract and compress different global features, thereby adapting to various learning tasks. The overall structure of the MambaLoc model is shown in Figure \ref{fig:MambaNonLocal} 

\textbf{1) Shared CNN Backbone for Compact Feature Extraction:} Given an image \( \mathbf{I} \in \mathbb{R}^{H\times W\times C} \), we employ the CNN Backbone at two different resolutions to generate activation maps \(\mathbf{ M_x} \in \mathbb{R}^{H\times W\times C_m}  \) and \( \mathbf{M_q} \in \mathbb{R}^{H\times W\times C_m}  \) for the tasks of position and orientation regression, respectively. To achieve positional encoding at the same depth, as suggested in \cite{carion2020end, shavit2021learning}, we apply a \( 1\times 1 \) convolution to linearly transform each activation map into a unified high-level depth dimension ($C_t=256$). Since the subsequent Transformer-encoder requires a sequence as input, we reshape the spatial dimensions of \( \mathbf{M} \in \mathbb{R}^{H_m \times W_m \times C_t} \) into a single dimension, resulting in \( \mathbf{\hat{M}} \in \mathbb{R}^{H_m \cdot W_m \times C_t} \) \cite{carion2020end}. To better distinguish the downstream orientation and position regression tasks, similar to the classification token in \cite{devlin2018bert}, we introduce a learnable token \( \mathbf{t} \in \mathbb{R}^{C_t} \) as the first token of each \( \mathbf{\widehat{M}} \). Consequently, the encoder input becomes \( \mathbf{E_\text{in}} = \left[ \mathbf{t}, \mathbf{\widehat{M}} \right] \in \mathbb{R}^{(H_m \cdot W_m + 1) \times C_t} \). Additionally, we adopt the positional encoding technique from \cite{shavit2021learning} to preserve the spatial information of each map location and assign unique positional encodings to the tokens. To reduce the number of learned positional parameters, we train two separate one-dimensional encodings: \(\mathbf{ E_x} \in \mathbb{R}^{(W_m+1)\times C_t /2} \) for the X-axis and \( \mathbf{E_y} \in \mathbb{R}^{(H_m+1)\times C_t /2} \) for the Y-axis. Thus, the positional embedding vectors for a 2-D spatial position \( (i,j) \), where \( i \in 1, \dotsc, H_m \) and \( j \in 1, \dotsc, W_m \), are represented as:

\begin{equation}
\mathbf{E}_{\text{pos}}^{i,j} = \begin{bmatrix} \mathbf{E}_x^j \\\mathbf{ E}_y^i \end{bmatrix} \in \mathbb{R}^{C_t}.
\end{equation}

These vectors are then reshaped into \( \mathbf{\widehat{E}} \in \mathbb{R}^{(H_m \cdot W_m + 1) \times C_t} \) before being fed into the subsequent Transformer Encoders.

\textbf{2) Separate Pose Regressing Transformer Encoders:} Following the approach of \cite{kendall2015posenet, shavit2024learning}, we employ a standard Transformer-Encoder architecture \cite{devlin2018bert} consisting of $n=6$ repeated blocks. Each block is composed of a multi-head self-attention (MHA) mechanism and a two-layer MLP with gelu activation. The inputs are normalized with LayerNorm \cite{vaswani2017attention} before each module, and residual connections along with dropout are used to combine the inputs with the outputs. As recommended by \cite{carion2020end}, positional encodings are added to the input before each layer, and an additional LayerNorm is applied to the final output. The encoder output (which also serves as the input for the subsequent GIS module), $\mathbf{G_{in}} \in \mathbb{R}^{C_t}$, at the special token $\mathbf{t}$, provides a comprehensive, context-aware summary of the local features from the input activation map.

\textbf{3) Separate Global Information Selectors (GIS) for Global Feature Extraction:} Although the Transformer Encoder can autonomously learn key features for camera localization, neural networks trained on specific scenes may overfit due to insufficient global image context, leading to slower convergence and reduced robustness in sparse training scenarios. One solution is to use Non-local Neural Networks \cite{wang2018non} to capture long-range dependencies, but their self-attention mechanism is computationally expensive. To address this challenge, we propose Global Information Selectors (GIS), which leverage the strengths of Mamba by utilizing a set of learnable parameters to selectively compress input features into a smaller hidden state. This approach results in a more compact and effective global feature representation. Additionally, we integrate this process with Mamba's Hardware-aware Algorithm to further optimize training efficiency, allowing a single Mamba layer to function as a Non-local Neural Network, efficiently capturing global information and accelerating convergence.

Specifically, the process begins by concatenating the output of the visual encoder, denoted as \( \mathbf{G_{in}} \), with its flipped version, \( \mathbf{G_{flip}} \). This combined input, \( \mathbf{G_{concat}} \), is subsequently fed into a bidirectional single-layer Mamba Module, which selectively compresses the input while utilizing a hardware-aware algorithm. This approach yields a more compact hidden state output, denoted as \(\mathbf{ \widehat{G}} \). Specifically, for the position and orientation encoders, we obtain \( \mathbf{\widehat{G_x}} \) and \( \mathbf{\widehat{G_q}} \), respectively. Please refer to Section \ref{The Global Information Selector (GIS)} for more details.

\textbf{4) 6-DoF Camera Pose Regressor}

The pose regressor employs an MLP head with a single hidden layer and a gelu activation function to map the global features extracted by the two separate Global Information Selectors ($\mathbf{\widehat{G_x}}, \mathbf{\widehat{G_q}}$) to the camera's position and quaternion $[\hat{\mathbf{x}}, \hat{\mathbf{q}}]$ (Equation \ref{eq:2}). The hidden layer in the MLP head expands the dimensionality from 256 to 1024 before passing the output vector to a fully connected layer for regression. During training, neural network parameters are optimised using the Euclidean distance (L2 Loss) for training images \( \mathbf{I} \) and their pose labels \( \mathbf{p}= [\mathbf{x}, \mathbf{q}] \). The loss function balances the position and rotation losses with the weights \( \beta \) and \( \gamma \) (Equation \ref{eq:3}), which  are learnt simultaneously during training with initial values \( \beta_0 \) and \( \gamma_0 \) by modeling the uncertainty of different tasks  \cite{kendall2017geometric}.

\begin{equation}
\label{eq:2}
[\hat{\mathbf{x}}, \hat{\mathbf{q}}] = [\text{MLP}(\mathbf{\widehat{G_x}}), \text{MLP}(\mathbf{\widehat{G_q}})).
\end{equation}

\begin{equation}
\label{eq:3}
\text{loss}(\mathbf{I}) = \left \| \mathbf{x} - \hat{\mathbf{x}} \right \|_2 e^{-\beta} + \beta + \left \| {\mathbf{q}} - \frac{\hat{\mathbf{q}}}{\left \| {\hat{\mathbf{q}}} \right \| 
 }  \right \|_2 e^{-\gamma} + \gamma.
\end{equation}

Note that in camera pose regression tasks, the quaternion $\mathbf{\hat{q}}$ is chosen for its ease of continuous and differentiable formulation in representing orientation \cite{kendall2017geometric}. To ensure mapping it to a valid rotation matrix, it is normalized to a unit vector $\frac{\hat{\mathbf{q}}}{\left \| {\hat{\mathbf{q}}} \right \| 
 } $ \cite{shavit2024learning}.

\subsection{The Global Information Selector (GIS)}
\label{The Global Information Selector (GIS)}
 Traditional Non-local Neural Networks, proposed by \cite{wang2018non}, capture long-range dependencies by computing the response at a position as a weighted sum of features at all positions in the input. The significant advantage of this approach is its ability to capture wide-range dependencies directly with few layers by calculating associations between any two positions. However, it also increases computational complexity compared to local operations.

 In this section, we will provide a detailed explanation of how we innovatively introduced the mechanism of Non-local Neural Networks into the Mamba model. The overall structure of the MambaLoc model is shown in Figure \ref{fig:MambaNonLocal}. Unlike traditional Non-local operations that compute attention across all elements without compression, we leverage Mamba's strengths to selectively compress input features into a smaller hidden state using a set of learnable parameters. This approach yields a more compact and effective global feature representation. Additionally, the process is further enhanced by integrating the Hardware-aware Algorithm to optimize training efficiency. This allows the model to capture global information while rapidly enhancing convergence capabilities. 
 
Let \( B \) represent the batch size, \( D \) the input feature channels, and \( N \) the state dimension of Mamba, set to 16 in all experiments. In the GIS module of MambaLoc, the input feature dimension is \( D = C_t \). The process flow is illustrated in Algorithm \ref{alg:MambaNonLocal},where:

\begin{algorithm}[H]
\caption{Global Information Selector (GIS)}
\label{alg:MambaNonLocal}
\begin{algorithmic}[1]
\REQUIRE $\mathbf{G_{in}} \in \mathbb{R}^{B \times 1 \times D}$ 
\ENSURE $\mathbf{\widehat{G}} \in \mathbb{R}^{B \times 1 \times D}$
\STATE \textbf{Flip and concatenate:} \\ $\mathbf{G_{flip}} \leftarrow \text{flip}(\mathbf{G_{in}}, \text{dim}=2)$
\\ $\mathbf{G_{concat}} \in \mathbb{R}^{B \times 2 \times D} \leftarrow \text{concatenate}(\mathbf{G_{flip}}, \mathbf{G_{in}}, \text{dim}=1)$
\STATE $\mathbf{A} \in \mathbb{R}^{D \times N} \leftarrow \text{Parameter}$
\\$\qquad\quad\textcolor{orange}{{\tiny \triangleright}}$\COMMENT{Represents structured $N \times N$ matrix}
\STATE $\mathbf{B} \in \mathbb{R}^{\textcolor{brown}{B \times 2 \times N}} \leftarrow \textcolor{brown}{s_B(\mathbf{G_{concat})}}$
\STATE $\mathbf{C} \in \mathbb{R}^{\textcolor{brown}{B \times 2 \times N}} \leftarrow \textcolor{brown}{\mathbf{s_C(G_{concat})}}$
\STATE $\mathbf{\Delta} \in \mathbb{R}^{\textcolor{brown}{B \times 2 \times D}} \leftarrow \tau_{\Delta}(\text{Parameter} + \textcolor{brown}{s_{\Delta}(\mathbf{G_{concat})}})$
\STATE $\mathbf{\overline{A} , \overline{B}}  \in \mathbb{R}^{\textcolor{brown}{B \times 2 \times D \times N}} \leftarrow \text{discretize}(\mathbf{\Delta, A, B})$
\STATE $\mathbf{\widehat{G}_{concat}} \leftarrow \text{SSM}(\mathbf{\overline{A}, \overline{B}, C)}(\mathbf{\widehat{G}_{concat}})$ \\ $\qquad\quad\textcolor{orange}{{\tiny \triangleright}}$\COMMENT{$\mathbf{\widehat{G}_{concat}}\in \mathbb{R}^{B \times 2 \times D}$ \\  \qquad\qquad \quad= $\text{concatenate}(\mathbf{\widehat{G}_\text{flip}, \widehat{G}}, \text{dim}=1)$}
\\$\qquad\quad\textcolor{orange}{{\tiny \triangleright}}$\COMMENT{\textcolor{brown}{Time-varying}: recurrence (\textcolor{brown}{scan}) only}
\RETURN $\mathbf{\widehat{G}}$
\end{algorithmic}
\end{algorithm}

\begin{enumerate}
    \item Unlike traditional Mamba, which is designed for modalities such as language, our task focuses primarily on processing visual data. To enhance the module's spatial perception capabilities for visual sequences, our module innovatively employs bidirectional sequence modeling. Specifically, after deriving one-dimensional features from visual encoders, we concatenate these features with their flipped counterparts to create a compact 2-D feature vector. This vector is then processed by a bidirectional single-layer Mamba Module, which compresses it into a smaller hidden state while incorporating a hardware-aware algorithm. This approach significantly improves the efficiency and effectiveness of handling visual data for our specific tasks.
    
    \item To ensure matrix $\mathbf{A}$ selectively retains feature information, we initialize it using the Hippo matrix \cite{fu2022hungry} derived from tracking the coefficients of a Legendre polynomial \cite{voelker2019legendre}, instead of random initialization.
    
    \item To enhance content awareness, the one-dimensional feature map \(\mathbf{G_{in}} \in \mathbb{R}^{B \times 1 \times D}\) is utilized to derive the input-dependent learnable parameters \(\mathbf{B}\), \(\mathbf{C}\), and \(\mathbf{\Delta}\) through three linear projections: \(s_B\), \(s_C\), and \(\tau_{\Delta}\). In this framework, \(\mathbf{B}\) and \(\mathbf{\Delta}\) capture "how the input influences the state," while \(\mathbf{C}\) captures "how the current state translates to the output." According to the zero-order hold method, \(\mathbf{\Delta}\) determines the resolution of the input during discretization. As a learnable variable, smaller elements in \(\mathbf{\Delta}\) tend to prioritize global information by de-emphasizing specific details, whereas larger elements in \(\mathbf{\Delta}\) focus more on specific details rather than surrounding information. Notably, \(\mathbf{\Delta}\) can dynamically adjust its dimension based on the input feature map's size. In our experiments, we set \(\mathbf{\Delta}\) to be one-sixteenth of the input feature dimension, D.

    \item Utilizing a gating mechanism similar to those in other sequence models (e.g., LSTM, GRU), we regulate and control the flow of information, specifically the selective scan output at different time steps. In this approach, the input feature \(\mathbf{G_{concat}}\) is linearly projected and then split into two equal parts: the hidden states and the gating signal. The gating signal is trained to determine the extent to which information from the current input feature map should be transmitted to the output at each time step, thereby enabling the model to more effectively capture the key features of the input image. Finally, the gating signal undergoes a non-linear activation function (SiLU) and is element-wise multiplied with the selective scan output.

\end{enumerate}

In line with \cite{gu2023mamba}, we adopt \(s_B(x) = \text{Linear}_N(x)\), \(s_C(x) = \text{Linear}_N(x)\), \(s_{\Delta}(x) = \text{Broadcast}_D(\text{Linear}_1(x))\), and \(\tau_{\Delta} = \text{softplus}\). Section \ref{Experiments} offers qualitative and quantitative comparisons of models using our bidirectional Global Information Selector (GIS) against those with traditional Mamba blocks and unprocessed feature maps, focusing on training speed and robustness to sparse viewpoints.

\subsection{MambaLoc Application on Terminal Devices}
\label{MambaLoc Application on Terminal Devices}
While MambaLoc is capable of operating on edge devices, it faces limitations, including dependency on the computational power of these devices and the requirement for stable network connections. Furthermore, the process of transferring data to and from edge devices, coupled with the computation and retrieval of results, can introduce latency, potentially affecting real-time performance. To mitigate these challenges, deploying the localization network directly on terminal devices—such as autonomous mobile robots, unmanned vehicles, and smartphones—presents a more efficient and robust solution. By doing so, the constraints of edge device configurations and network dependencies can be eliminated, while enabling faster local image processing, optimizing processing efficiency, and reducing energy consumption, thus facilitating the broader practical deployment of the network. Therefore, this section explores the design considerations for deploying MambaLoc on terminal devices.

However, directly deploying existing deep learning-based camera localization networks on terminal devices presents significant challenges due to their typically large number of parameters, which makes them impractical for hardware-constrained mobile environments. To overcome this issue while maintaining accuracy, we adopted a hybrid knowledge distillation approach that integrates feature-based and logits-based techniques. This approach allowed us to distill a complex network trained on edge devices into a more efficient student network suitable for deployment on terminal devices. 

This section outlines the methodology of our hybrid knowledge distillation approach for deploying MambaLoc on terminal devices. This approach emphasizes the optimization of model parameters to ensure efficient deployment and optimal performance. To achieve a balance between latency and accuracy, we selected EfficientNet-B0 \cite{tan2019efficientnet}, a compact network that optimizes computational efficiency and performance, as the backbone of our student model. Features extracted by EfficientNet-B0 are subjected to adaptive average pooling and dropout to mitigate overfitting, followed by processing through three linear layers. The first layer reduces the 1280-dimensional feature vector to a 1024-dimensional latent space, facilitating the extraction of salient features for pose regression. The final output is a 7-dimensional vector, representing the camera's pose in terms of translation and quaternion rotation components. During the offline distillation process, the weights of the pre-trained teacher model were kept frozen. To enhance feature extraction learning from the teacher network, we employed an offline distillation approach, utilizing the distilled loss function as specified in Equation \eqref{eq:6}:

\begin{equation}
\label{eq:6}
    \mathcal{L}_D = \mathcal{L}_L + \mathcal{L}_S + \mathcal{L}_F.
\end{equation}
,where \( \mathcal{L}_L \), \( \mathcal{L}_S \), and \( \mathcal{L}_F \) represent the logits-based hard distillation loss, the logits-based soft distillation loss, and the feature-based distillation loss, respectively. The form of \( \mathcal{L}_L \) is identical to that in Equation \eqref{eq:3}, while the details of \( \mathcal{L}_S \) and \( \mathcal{L}_F \) are provided in the following paragraphs.

\textbf{Logits Based Soft-Distillation Loss} 
The Logits-Based Soft-Distillation Loss, defined in Equation \ref{eq:5}, is applied to the estimated poses from the student and teacher networks, denoted as \(\mathbf{p_s} = [\mathbf{x_s}, \mathbf{q_s}]\) and \(\mathbf{p_t} = [\mathbf{x_t}, \mathbf{q_t}]\), respectively. This loss function balances the position and orientation losses by modeling the uncertainty of each task through the weights \(\beta_{soft}\) and \(\gamma_{soft}\), as proposed by Kendall et al.~\cite{kendall2017geometric}. These weights, initially set to \(\beta_{soft0}\) and \(\gamma_{soft0}\), are learned adaptively during training to optimize the distillation process.

\begin{equation}
\label{eq:5}
\mathcal{L}_S = \mathcal{L}_{\text{KL}}(\hat{\mathbf{x_s}}, \hat{\mathbf{x_t}}) e^{-\beta_{soft}} + \beta_{soft} + 
\mathcal{L}_{\text{KL}}(\hat{\mathbf{q_s}}, \hat{\mathbf{q_t}}) e^{-\gamma_{soft}} + \gamma_{soft}.
\end{equation}

where the Kullback-Leibler loss $\mathcal{L}_{\text{KL}}$ is defined as:

\begin{align}
\label{eq:8}
\mathcal{L}_{\text{KL}}=\mathcal{L}_{\text{KD}}(\mathcal{F_\text{LogS}}(T^{-1} \cdot \hat{\mathbf{y_s}}),\mathcal{F_{S}}(T^{-1} \cdot \hat{\mathbf{y_t}})) \cdot T^2
\end{align}

Here, $\hat{\mathbf{y}}$ may represent either $\hat{\mathbf{x}}$ for position or $\hat{\mathbf{q}}$ for orientation. The temperature parameter $T$ is set to 10.0 for all experiments, and $\mathcal{L}_{\text{KD}}$ denotes the Kullback-Leibler Divergence loss function. The functions $\mathcal{F}_{\text{LogS}}$ and $\mathcal{F}_{\text{S}}$ correspond to the $\mathbf{ \text{LogSoftmax}}$ and $\text{Softmax}$ functions, respectively.

\textbf{Feature-Based Distillation Loss} 
\begin{equation}
\label{eq:10}
\mathcal{L}_F = 1 - \text{cos}(\text{ft}, \text{fs}).\text{mean()}
\end{equation}

The feature-based distillation loss function \( \mathcal{L}_F \), described in Equation \eqref{eq:10}, aims to encourage similarity between flattened feature maps from the teacher model (\( \text{ft} \)) and the student model (\( \text{fs} \)). This loss computes the cosine similarity between these feature representations and then averages the results across all feature vectors.

	\begin{table*}[t]
		\centering
		\caption{Performance Evaluation of MambaLoc and Previous APR-Based Localization Methods on the 7-Scenes Dataset \cite{glocker2013real}: A Comparison of Pose Regression Results, Total Training Time in Terms of Median Translation Error (m), Rotation Error (°), Training Time (minutes). *For the method where the training time is not mentioned in the paper and the code is not open-sourced, we used "--" as a placeholder.}
        \label{tab:Edge_Device_7scenes_main_table}
		\tabcolsep 5pt	
		\begin{tabular}{@{}c|ccccccccc@{}}
			\toprule
			\begin{tabular}[c]{@{}c@{}}Method/Scenes\\ (\#trainset)\\ (\#testset)\end{tabular} &
			\begin{tabular}[c]{@{}c@{}}Heads\\ \#1000\\ \#1000\end{tabular} &
			\begin{tabular}[c]{@{}c@{}}Fire\\ \#2000\\ \#2000\end{tabular} &
			\begin{tabular}[c]{@{}c@{}}Kitchen\\ \#7000\\ \#1000\end{tabular} &
			\begin{tabular}[c]{@{}c@{}}Chess\\ \#4000\\ \#2000\end{tabular} &
			\begin{tabular}[c]{@{}c@{}}Office\\ \#6000\\ \#4000\end{tabular} &
			\begin{tabular}[c]{@{}c@{}}Pumpkin\\ \#4000\\ \#2000\end{tabular} &
			\begin{tabular}[c]{@{}c@{}}Stairs\\ \#2000\\ \#1000\end{tabular} &
			\begin{tabular}[c]{@{}c@{}}Avg.\\ Pose\\ Error\end{tabular} &
			\begin{tabular}[c]{@{}c@{}}Avg.\\ Training \\ Time\end{tabular} \\ \midrule
			\textbf{PoseNet(2015)\cite{kendall2015posenet}} &
			0.29/12.0 &
			0.47/14.4 &
			0.59/8.64 &
			0.32/8.12 &
			0.48/7.68 &
			0.47/8.42 &
			0.47/13.8 &
			0.44/10.4 &
			197 \\
			\textbf{PN Learn $\sigma$\textasciicircum{}2 (2017) \cite{kendall2017geometric}} &
			0.18/12.1 &
			0.27/11.8 &
			0.24/5.52 &
			0.14/4.50 &
			0.20/5.77 &
			0.25/4.82 &
			0.37/10.6 &
			0.24/7.87 &
			-- \\
			\textbf{Geo.PN (2017) \cite{kendall2017geometric}} &
			0.17/13.0 &
			0.27/11.3 &
			0.23/5.35 &
			0.13/4.48 &
			0.19/5.55 &
			0.26/4.75 &
			0.35/12.4 &
			0.23/8.12 &
			-- \\
			\textbf{LSTM PN (2017) \cite{walch2017image}} &
			0.21/13.7 &
			0.34/11.9 &
			0.37/8.83 &
			0.24/5.77 &
			0.30/8.08 &
			0.33/7.00 &
			0.40/13.7 &
			0.31/9.85 &
			60 \\
			\textbf{MapNet (2018) \cite{kendall2017geometric}} &
			0.18/13.3 &
			0.27/11.7 &
			0.23/\textbf{4.93} &
			\textbf{0.08/3.25} &
			\textbf{0.17}/5.15 &
			0.22/\textbf{4.02} &
			0.30/12.1 &
			0.21/7.77 &
			99 \\
			\textbf{AtLoc (2019) \cite{wang2020atloc}} &
			0.16/\textbf{11.8} &
			0.25/11.4 &
			0.23/5.42 &
			0.10/4.07 &
			\textbf{0.17}/5.34 &
			0.21/4.37 &
			0.26/10.50 &
			0.20/\textbf{7.56} &
			109 \\
			
			\textbf{\begin{tabular}[c]{@{}c@{}} TransPoseNet (2024)\cite{shavit2024learning}\end{tabular}} &
			\textbf{0.13}/12.7 &
			\textbf{0.24}/\textbf{10.6} &
			0.19/6.75 &
			\textbf{0.08}/5.68 &
			\textbf{0.17}/6.34 &
			0.17/5.60 &
			0.30/\textbf{7.02} &
			0.18/7.78&
			178 \\
			\textbf{\begin{tabular}[c]{@{}c@{}}TransBoNet (2024) \cite{song2024transbonet}\end{tabular}} &
			0.18/14.00 &
			0.25/12.46 &
			\textbf{0.17}/5.35 &
			0.11/4.48 &
			0.20/\textbf{5.08} &
			0.19/4.77 &
			0.30/13.04 &
			0.20/8.45 &
			40 \\
			\textbf{\begin{tabular}[c]{@{}c@{}}MambaLoc (ours)\end{tabular}} &
			\begin{tabular}[c]{@{}c@{}}\textbf{0.13}/12.30 \end{tabular} &
			\begin{tabular}[c]{@{}c@{}}0.25/10.79\end{tabular} &
			\begin{tabular}[c]{@{}c@{}}\textbf{0.17}/7.85\end{tabular} &
			\begin{tabular}[c]{@{}c@{}}0.09/6.14\end{tabular} &
			\begin{tabular}[c]{@{}c@{}}\textbf{0.17}/6.58\end{tabular} &
			\begin{tabular}[c]{@{}c@{}}\textbf{0.16}/6.56\end{tabular} &
			\begin{tabular}[c]{@{}c@{}}\textbf{0.23}/10.74\end{tabular} &
			\begin{tabular}[c]{@{}c@{}}\textbf{0.17}/8.71\end{tabular} &
			\begin{tabular}[c]{@{}c@{}}\textbf{14}\end{tabular}  \\

            \bottomrule
		\end{tabular}
	\end{table*}

\begin{table*}[h]
   \caption{Performance Evaluation of MambaLoc and Previous APR-Based Localization Methods on the Cambridge Landmark Dataset \cite{kendall2015posenet}: A Comparison of Pose Regression Results, Total Training Time in Terms of Median Translation Error (m), Rotation Error (°), Training Time (minutes). *For the method where the training time is not mentioned in the paper and the code is not open-sourced, we used "--" as a placeholder.}
    \label{tab:Edge_Device_Cambridge}
    \centering
    \begin{tabular}{l|c|c|c|c|c|c}
        \hline
        \textbf{Methods} & \textbf{Kings} & \textbf{Hospital} & \textbf{Shop} & \textbf{Church} & \textbf{Average} & \textbf{Avg. Training Time} \\
        \hline
        PoseNet (2015)\cite{kendall2015posenet} & 1.66/4.86 & 2.62/4.90 & 1.41/7.18 & 2.45/7.96 & 2.04/6.23 & 51 \\
        
       PN Learn $\sigma$\textasciicircum{}2 (2017) \cite{kendall2017geometric} & 0.99/1.06 & 2.17/\textbf{2.94} &1.05/3.97 & 1.49/3.43& 1.43/\textbf{2.85} & -- \\
        geo. PN (2017) \cite{kendall2017geometric} & 0.88/\textbf{1.04} & 3.20/3.29 & 0.88/3.78 & 1.57/\textbf{3.32} & 1.63/2.86 & -- \\
        LSTM PN (2017)\cite{walch2017image} & 0.99/3.65 & 1.51/4.29 & 1.18/7.44 & 1.52/6.68 & 1.30/5.51 & 95 \\
        
        MapNet (2018) \cite{kendall2017geometric} & 1.07/1.89 & 1.94/3.91 & 1.49/4.22 & 2.00/4.53 & 1.63/3.64 & 50 \\
        TransPoseNet (2024)\cite{shavit2024learning} & 0.60/2.43 & \textbf{1.45}/3.08 & 0.55/\textbf{3.49} & 1.09/4.99 &\textbf{ 0.91}/3.50 & 88 \\
        
        
        MambaLoc (ours) & \textbf{0.58}/2.64 & 1.59/2.95 & \textbf{0.53}/4.16 &\textbf{ 1.08}/5.61 & 0.95/3.84 & \textbf{38} \\
        \hline
    \end{tabular}
\end{table*}

\begin{table}[t]
    \centering
    \caption{Detailed Total Training Time (Minutes) for Each Scene in the 7-Scenes Dataset \cite{glocker2013real} Using MambaLoc.}
    \label{tab:Edge_Device_7scenes_time}
    \resizebox{0.5\textwidth}{!}{
    \begin{tabular}{@{}c|cccccccc@{}}
        \toprule
        \begin{tabular}[c]{@{}c@{}}Method/Scenes\\ (\#trainset)\\ (\#testset)\end{tabular} &
        \begin{tabular}[c]{@{}c@{}}Pumpkin\\ \#4000\\ \#2000\end{tabular} &
        \begin{tabular}[c]{@{}c@{}}Stairs\\ \#2000\\ \#1000\end{tabular} &
        \begin{tabular}[c]{@{}c@{}}Heads\\ \#1000\\ \#1000\end{tabular} &
        \begin{tabular}[c]{@{}c@{}}Fire\\ \#2000\\ \#2000\end{tabular} &
        \begin{tabular}[c]{@{}c@{}}Chess\\ \#4000\\ \#2000\end{tabular} &
        \begin{tabular}[c]{@{}c@{}}Office\\ \#6000\\ \#4000\end{tabular} &
        \begin{tabular}[c]{@{}c@{}}Kitchen\\ \#7000\\ \#1000\end{tabular} &
        \begin{tabular}[c]{@{}c@{}}Avg.\\ Training \\ Time\end{tabular} \\ \midrule
        MambaLoc &
        15 &
        3 &
        2 &
        4 &
        29 &
        27 &
        19 &
        14 \\
        \bottomrule
    \end{tabular}
    }
\end{table}

\begin{table}[t]
    \centering
    \caption{Detailed Total Training Time (Minutes) for Each Scene in the Cambridge Landmark Dataset \cite{kendall2015posenet} Using MambaLoc.}
    \label{tab:Edge_Device_cam_time}
    \resizebox{0.5\textwidth}{!}{
    \begin{tabular}{@{}c|cccccc@{}}
        \toprule
        \begin{tabular}[c]{@{}c@{}}Method/Scenes\\ (\#trainset)\\ (\#testset)\end{tabular} &
        \begin{tabular}[c]{@{}c@{}}KingsCollege\\ \#1220\\ \#343\end{tabular} &
        \begin{tabular}[c]{@{}c@{}}OldHospital\\ \#895\\ \#182\end{tabular} &
        \begin{tabular}[c]{@{}c@{}}ShopFacade\\ \#231\\ \#103\end{tabular} &
        \begin{tabular}[c]{@{}c@{}}StMarysChurch\\ \#1487\\ \#530\end{tabular} &
        \begin{tabular}[c]{@{}c@{}}Avg.\\ Training \\ Time\end{tabular} \\ \midrule
        MambaLoc &
        74 &
        23 &
        32 &
        23 &
        38 \\
        \bottomrule
    \end{tabular}
    }
\end{table}

\section{Experiments}
\label{Experiments}

The structure of this section is as follows: We first evaluate the accuracy and training efficiency of our method using standard indoor and outdoor benchmarks for camera pose estimation, comparing it with recent state-of-the-art localization techniques to demonstrate its effectiveness. Subsequently, we explore the generalization capability of the Global Information Selector (GIS) across various visual localization models, evaluating how well our method adapts to different model architectures. We then address the robustness of our approach by analyzing its performance with sparse training data, which is crucial for applications that require efficient utilization of limited data. Following this, we present the performance of our method on terminal devices, showcasing its versatility across different deployment scenarios. Finally, we conducted an ablation study to compare the optimization effects of the Global Information Selector (GIS) on MambaLoc against configurations using the traditional Mamba and without any Mamba. This study qualitatively and quantitatively demonstrates the effectiveness of GIS in enhancing training speed.

\subsection{Datasets}

\textbf{The Cambridge Landmarks dataset} \cite{kendall2015posenet} features urban landscapes for outdoor localization tasks within specified spatial ranges, with scene areas ranging from 875 to 5600 square meters. Each scene includes between 200 and 1500 training samples. For our comparative evaluation, similar to \cite{chen2021direct, chen2022dfnet, shavit2024learning, song2024transbonet}, we selected four scenes from this dataset, excluding the other two as they are less commonly used for benchmarking purposes. \textbf{The 7Scenes dataset} \cite{glocker2013real} comprises seven small-scale indoor environments, each covering a distinct spatial area ranging from 1 to 18 cubic meters. The scenes contain between 1,000 and 7,000 training samples and 1,000 to 5,000 validation samples. Both datasets pose various localization challenges, including occlusions, reflections, motion blur, varying lighting conditions, repetitive textures, and changes in viewpoints and trajectories.

\subsection{Implementation Details}

Our models are implemented in PyTorch and trained using the Adam optimizer with hyperparameters $\beta_1=0.9$, $\beta_2=0.999$, $\epsilon=1\text{e}^{-10}$, and a batch size of 8. The learning rate is set to $\lambda=1\text{e}^{-4}$ and decreases by a factor of 10 every 100 epochs for indoor localization (or every 200 epochs for outdoor localization). Training continues for up to 600 epochs unless an early stopping criterion is met. This criterion is based on the validation set, where the validation loss is monitored each epoch, and early stopping is triggered if the loss does not decrease from the previous low for more than five consecutive epochs. We apply a weight decay of $1\text{e}^{-4}$ and a dropout rate of 0.1 to train the encoders. For pose estimation tasks, parameters are initialized with $\beta_0 = -0.5$ and $\gamma_0 = -6.5$. During the distillation procedure, parameters are initialized with $\beta_{\text{soft0}} = -0.5$ and $\gamma_{\text{soft0}} = -6.5$. To improve the model's generalization ability, we use the data augmentation method proposed by \cite{kendall2015posenet}. During training, images are resized so that their shorter edge measures 256 pixels, followed by a random crop to 224 × 224 pixels. Additionally, we apply random adjustments to brightness, contrast, and saturation. During testing, images are resized and a center crop is taken without further augmentations. To validate the efficiency of MambaLoc's training process, we conducted a comparative analysis of training time between our model and previously published open-source methods. For consistency, all training durations were measured using the same NVIDIA L20 server. 

It is important to note that this section focuses on evaluating our model's ability to achieve state-of-the-art (SOTA) accuracy in the shortest possible time. Therefore, we do not compare our method with multi-scene approaches (e.g., \cite{shavit2021learning,shavit2023coarse}) or methods that incorporate additional NeRF training (e.g., \cite{chen2021direct,chen2022dfnet}). Additionally, all experimental results for MambaLoc presented here are obtained from a single training phase, without any secondary fine-tuning stages (e.g., TransPoseNet \cite{shavit2024learning}).

\subsection{Comparative Analysis of the Accuracy and Training Time of Visual Localization Models}

\subsubsection{Indoor Localization}

\begin{figure*}[h]
    \centering
    \caption{\textbf{Camera Localization Results on the 7-Scenes dataset \cite{glocker2013real}.} Each subfigure contains a 3D plot at the top, illustrating the camera trajectories (green for ground truth, red for predicted), and a color bar at the bottom representing the rotation error for all frames. Refer to Table \ref{tab:Edge_Device_7scenes_main_table} for a quantitative comparison.
}
    \label{fig:2dmap_7scenes}
    \includegraphics[width=\textwidth]{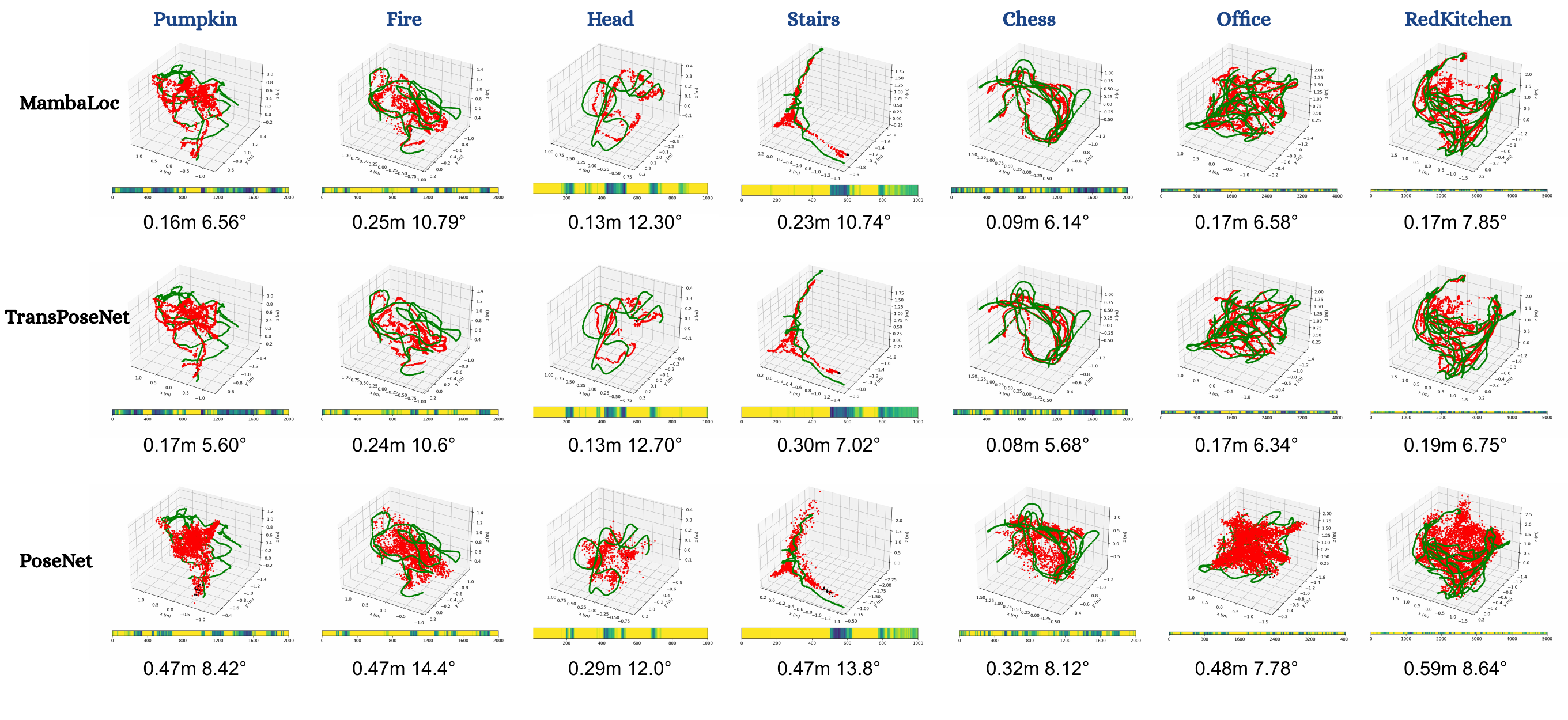}

\end{figure*}

We evaluated the performance of our method in terms of pose estimation accuracy using the 7-Scenes indoor camera localization dataset \cite{glocker2013real}. Figure \ref{fig:2dmap_7scenes} and Table \ref{tab:Edge_Device_7scenes_main_table} provides a quantitative comparison of the estimated trajectories produced by our method and those of previous approaches. MambaLoc achieves state-of-the-art (SOTA) accuracy in average translation accuracy, with results that are on par with the best-performing methods. Notably, this level of accuracy is achieved in a single training phase, unlike methods such as TransPoseNet \cite{shavit2024learning}, which require additional fine-tuning stages.

In addition to accuracy, we assessed the training efficiency of different methods by comparing their training durations. Table \ref{tab:Edge_Device_7scenes_main_table} presents a detailed comparison of our method with earlier single-frame absolute pose regression (APR) methods. To ensure fairness, we measured the training durations of previously open-source methods on the same server (NVIDIA L20). Our approach significantly reduces the average training duration—by 65\% compared to the fastest method (TransBoNet \cite{song2024transbonet}) and by 92.9\% compared to the slowest method (PoseNet \cite{kendall2015posenet}). Detailed total training durations for each scene in the 7-Scenes dataset \cite{glocker2013real} using MambaLoc are provided in Table \ref{tab:Edge_Device_7scenes_time}. For the Heads scene, which contains 1,000 images, our method completes training in just 2 minutes, while still achieving SOTA accuracy.

These results collectively demonstrate that our approach not only achieves high accuracy but also enables rapid training on edge devices, making end-to-end localization models feasible for productization and practical applications.

\subsubsection{Outdoor Localization}

\begin{figure*}[h]
    \centering
    \caption{\textbf{Camera Localization Results on the Cambridge Landmark dataset \cite{kendall2015posenet}.} The black line represents the ground truth trajectory, with the star marking the first frame, while the red lines depict the predicted camera poses. Each figure's caption indicates the mean translation error (m) and mean rotation error (°). Refer to Table \ref{tab:Edge_Device_Cambridge} for a quantitative comparison.
}
    \includegraphics[width=\textwidth]{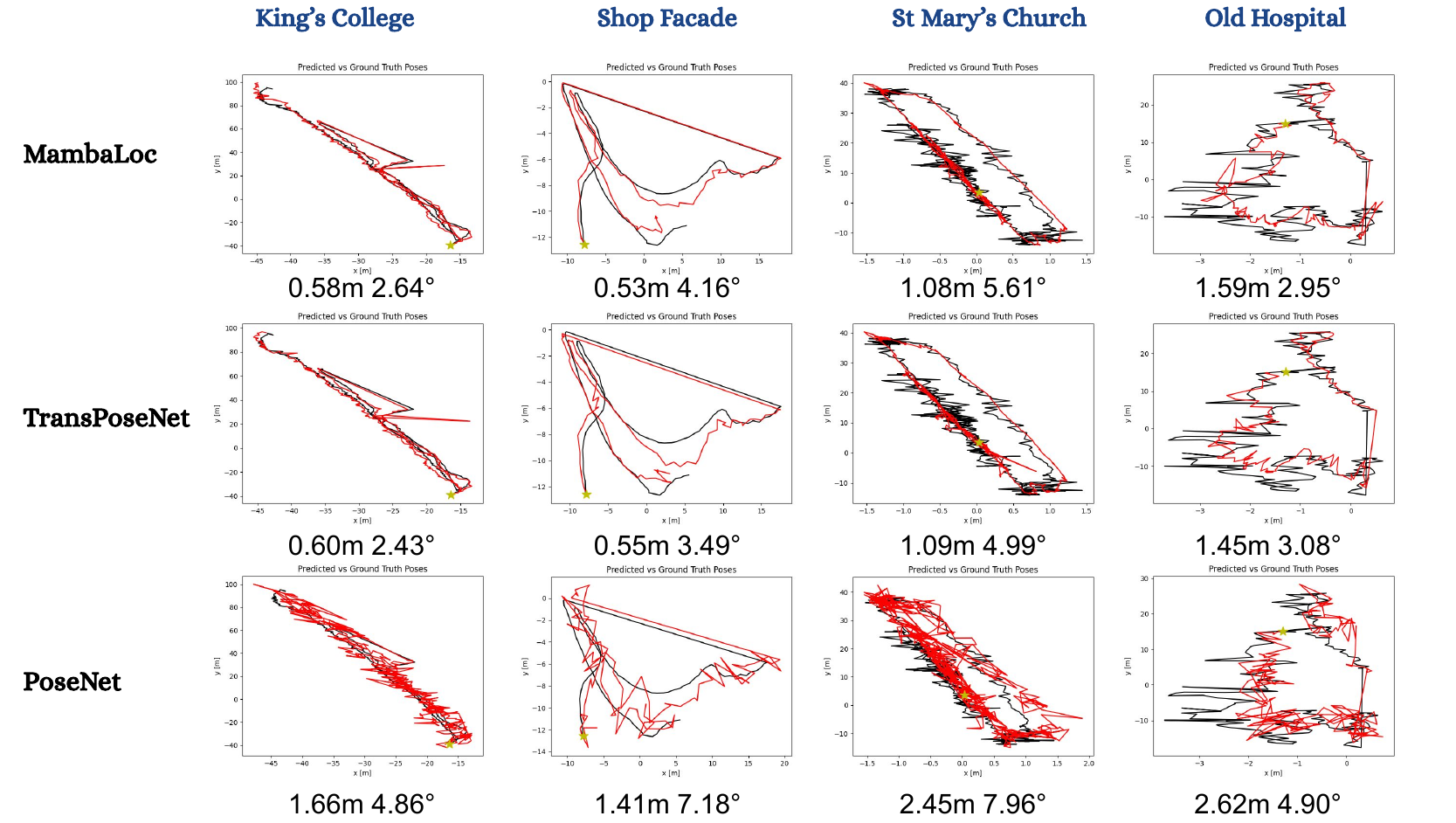}

    \label{fig:cam_2d_map}
\end{figure*}

We further evaluated our approach on four outdoor scenes from the Cambridge Landmarks dataset \cite{kendall2015posenet}. Figure \ref{fig:cam_2d_map} provides a quantitative comparison of the estimated trajectories produced by our method versus those from previous approaches. Table \ref{tab:Edge_Device_Cambridge} presents a comparison of our method's accuracy and training time against other APR methods, excluding those that did not report results on the Cambridge dataset. All training durations were measured on the same NVIDIA L20 server to ensure consistency. 

Our findings indicate that, even on larger and more challenging outdoor datasets, our model demonstrates significantly greater training efficiency while still achieving pose estimation accuracy comparable to state-of-the-art (SOTA) methods. Importantly, this level of accuracy is achieved in a single training phase, without the need for additional fine-tuning stages (as required by TransPoseNet \cite{shavit2024learning}).

Regarding the training duration, our method reduces the average training duration by 24\% compared to the fastest method (MapNet \cite{kendall2017geometric}) and by 60\% compared to the slowest method (LSTM PN \cite{walch2017image}). Detailed total training duration for each scene in the Cambridge dataset \cite{kendall2015posenet} using MambaLoc are provided in Table \ref{tab:Edge_Device_cam_time}. These results underscore our approach's effectiveness in balancing training duration and accuracy, making it a highly efficient solution for real-world applications.

\begin{table*}[h!]
		\centering
		\caption{Performance Evaluation of the Generalization Capability of the Global Information Selector (GIS) Across Various Visual Localization Models on the 7-Scenes Dataset \cite{glocker2013real}: A Comparison of Pose Regression Results, Total Training Time, and Average Inference Time in Terms of Median Translation Error (m), Rotation Error (°), Training Time (minutes), and Inference Time per Frame (milliseconds)}
		\label{tab:Generalization}
		\begin{tabular}{|c|cc|cc|cc|}
			\hline
			\multirow{2}{*}{\textbf{Methods}} &
			\multicolumn{2}{c|}{\textbf{Pose   Error}} &
			\multicolumn{2}{c|}{\textbf{Avg. Training   Time (min)}} &
			\multicolumn{2}{c|}{\textbf{Avg. Inference   Time/Frame (milliseconds)}} \\ \cline{2-7} 
			&
			\multicolumn{1}{c|}{\textbf{Benchmark}} &
			\textbf{\begin{tabular}[c]{@{}c@{}}Benchmark \\ w. GIS\end{tabular}} &
			\multicolumn{1}{c|}{\textbf{Benchmark}} &
			\textbf{\begin{tabular}[c]{@{}c@{}}Benchmark \\ w. GIS\end{tabular}} &
			\multicolumn{1}{c|}{\textbf{Benchmark}} &
			\textbf{\begin{tabular}[c]{@{}c@{}}Benchmark \\ w. GIS\end{tabular}} \\ \hline
			\textbf{PoseNet\cite{kendall2015posenet}} &
			\multicolumn{1}{c|}{0.44/10.4} &
		0.37/8.99 &
			\multicolumn{1}{c|}{197} & 23
			&
			\multicolumn{1}{c|}{\textbf{2.56}} & 3.09
			\\ \hline
			\textbf{LSTM PN \cite{walch2017image}} &
			\multicolumn{1}{c|}{0.31/9.85} &
			0.29/8.93 &
			\multicolumn{1}{c|}{60} & 18
			& 
			\multicolumn{1}{c|}{9.00} & 9.00
			\\ \hline
			\textbf{TransPoseNet \cite{shavit2024learning}} &
			\multicolumn{1}{c|}{0.18/\textbf{7.78}} &
			\textbf{0.17}/8.71 &
			\multicolumn{1}{c|}{178} &
			\textbf{14} &
			\multicolumn{1}{c|}{10.64} & 11.80
			\\ \hline
		\end{tabular}
	\end{table*}

	\begin{table}[h!]
		\centering
		\caption{Comparison of Model Sizes for Different Localization Models Before and After Adding the Global Information Selector (GIS): Evaluation Based on the Number of Learnable Parameters (in Mb)}
		\label{tab:model_size}
		\begin{tabular}{|c|c|c|}
			\hline
			Method & \begin{tabular}[c]{@{}c@{}}Benchmark\\     Model Size (Mb)\end{tabular} & \begin{tabular}[c]{@{}c@{}}Benchmark \\w. GIS\\     Model Size (Mb)\end{tabular} \\ \hline
			PoseNet \cite{kendall2015posenet}      & 85.26 & 186.12 \\ \hline
			AtLoc \cite{wang2020atloc}       & 93.3  & 118.69 \\ \hline
			LSTM PN \cite{walch2017image}     & 89.6  & 115.44 \\ \hline
			TransPoseNet \cite{shavit2024learning} & 40.57 & 42.24  \\ \hline
		\end{tabular}
	\end{table}

\begin{table*}[h!]
	\centering
	\caption{Evaluation of the Robustness Enhancement of Various Absolute Pose Regression (APR) Methods by the Global Information Selector (GIS) on Sparse-View Training Sets: A Comparative Performance Analysis of Visual Localization Models on the 7-Scenes Dataset \cite{glocker2013real} in Terms of Median Translation Error (m) and Rotation Error (°). We evaluated pose estimation errors after uniformly sampling 100\%, 10\%, and 5\% of the training set. The arrow direction indicates the increase/decrease in pose error of Benchmark-GIS relative to Benchmark under the same sparsity conditions.}
	\label{tab:NLMB_sparsity}
	\begin{tabular}{|c!{\vrule width 2pt}cc!{\vrule width 2pt}cc!{\vrule width 2pt}cc!{\vrule width 2pt}cc|}
		\hline
		\multirow{2}{*}{Methods} &
		\multicolumn{2}{c!{\vrule width 2pt}}{100\% Train Set} &
		\multicolumn{2}{c!{\vrule width 2pt}}{10\% Train Set} &
		\multicolumn{2}{c!{\vrule width 2pt}}{5\% Train Set} &
		\multicolumn{2}{c|}{\textbf{Avg. Pose Error}} \\ \cline{2-9} 
		&
		\multicolumn{1}{c|}{Benchmark} &
		\begin{tabular}[c]{@{}c@{}}Benchmark \\ w. GIS\end{tabular} &
		\multicolumn{1}{c|}{Benchmark} &
		\begin{tabular}[c]{@{}c@{}}Benchmark \\ w. GIS\end{tabular} &
		\multicolumn{1}{c|}{Benchmark} &
		\begin{tabular}[c]{@{}c@{}}Benchmark \\ w. GIS\end{tabular} &
		\multicolumn{1}{c|}{\textbf{Benchmark}} &
		\textbf{\begin{tabular}[c]{@{}c@{}}Benchmark \\ w. GIS\end{tabular}} \\ \hline
		PoseNet\cite{kendall2015posenet} &
		\multicolumn{1}{c|}{0.44/10.4} &
		$0.37\textcolor{red}{\downarrow}/8.99\textcolor{red}{\downarrow}$ &
		\multicolumn{1}{c|}{0.65/10.60} & $0.44\textcolor{red}{\downarrow}/10.25\textcolor{red}{\downarrow}$ &
		\multicolumn{1}{c|}{1.53/10.29} & $0.54\textcolor{red}{\downarrow}/10.65\textcolor{red}{\uparrow}$ &
		\multicolumn{1}{c|}{0.87/10.43} & \textbf{0.45/9.96} \\ \hline
		LSTM PN \cite{walch2017image} &
		\multicolumn{1}{c|}{0.31/9.85} &
		$0.29\textcolor{red}{\downarrow}/8.93\textcolor{red}{\downarrow}$ &
		\multicolumn{1}{c|}{2.97/10.32} & $0.25\textcolor{red}{\downarrow}/9.37\textcolor{red}{\downarrow}$ &
		\multicolumn{1}{c|}{3.65/11.30} & $0.27\textcolor{red}{\downarrow}/11.98\textcolor{red}{\uparrow}$ &
		\multicolumn{1}{c|}{2.31/10.49} & \textbf{0.27/10.09} \\ \hline
		TransPoseNet \cite{shavit2024learning} &
		\multicolumn{1}{c|}{0.18/7.78} &
		$0.17\textcolor{red}{\downarrow}/8.71\textcolor{red}{\uparrow}$ &
		\multicolumn{1}{c|}{0.21/8.04} &
		$0.19\textcolor{red}{\downarrow}/8.55\textcolor{red}{\uparrow}$ &
		\multicolumn{1}{c|}{0.21/9.12} & $0.21/8.87\textcolor{red}{\downarrow}$ &
		\multicolumn{1}{c|}{0.20/\textbf{8.31}} & \textbf{0.19}/8.71 \\ \hline
	\end{tabular}
\end{table*}

\begin{table*}[h!]
    \centering
    \caption{Performance Evaluation of MambaLoc and Previous APR-Based Localization Methods for Robustness to Different Levels of Sparse-View Training Scenarios on the 7-Scenes Dataset \cite{glocker2013real}: A Comparison of Pose Regression Results in Terms of Median Translation Error (m) and Rotation Error (°)}
    \label{tab:sparsity_performance_comparison}
    \small
    \begin{tabular}{l|c|c|c|c|c|c}
        \hline
        \textbf{Methods/Sparsity} & \textbf{1} & \textbf{1/2} & \textbf{1/3} & \textbf{1/10} & \textbf{1/20} & \textbf{Avg. Pose Error} \\
        \hline
        \textbf{PoseNet} \cite{kendall2015posenet}& \begin{tabular}[c]{@{}c@{}}0.44/10.40\end{tabular} & \begin{tabular}[c]{@{}c@{}}0.47/10.29\end{tabular} & \begin{tabular}[c]{@{}c@{}}0.52/9.89\end{tabular} & \begin{tabular}[c]{@{}c@{}}0.65/10.60\end{tabular} & \begin{tabular}[c]{@{}c@{}}1.53/10.29\end{tabular} &
        \begin{tabular}[c]{@{}c@{}}0.72/10.29 \end{tabular}
        
        \\
        \hline
        \textbf{TransPoseNet \cite{shavit2024learning}} & \begin{tabular}[c]{@{}c@{}}0.18/\textbf{7.78}\end{tabular} & \begin{tabular}[c]{@{}c@{}}0.24/\textbf{7.58}\end{tabular} & \begin{tabular}[c]{@{}c@{}}0.17/\textbf{8.19}\end{tabular} & \begin{tabular}[c]{@{}c@{}}0.21/\textbf{8.04}\end{tabular} & \begin{tabular}[c]{@{}c@{}}\textbf{0.21}/9.12 \end{tabular}  &
        \begin{tabular}[c]{@{}c@{}}0.20/\textbf{8.14} \end{tabular}
        
        \\
        \hline
        \textbf{MambaLoc} & \begin{tabular}[c]{@{}c@{}}\textbf{0.17}/8.71  \end{tabular} & \begin{tabular}[c]{@{}c@{}}\textbf{0.18}/8.54\end{tabular} & \begin{tabular}[c]{@{}c@{}}\textbf{0.16}/8.50\end{tabular} & \begin{tabular}[c]{@{}c@{}}\textbf{0.19}/8.55\end{tabular} & \begin{tabular}[c]{@{}c@{}}\textbf{0.21}/\textbf{8.87} \end{tabular}  &
        \begin{tabular}[c]{@{}c@{}}\textbf{0.18}/8.63 \end{tabular}
        
        \\
        \hline

    \end{tabular}
\end{table*}

\subsection{Evaluation of the Generalization Capability of the Global Information Selector (GIS) Across Various Visual Localization Models}

\textbf{Quantitative Evaluation:} To evaluate the effectiveness of the proposed Global Information Selector (GIS) as a versatile module for enhancing existing end-to-end localization models, we conducted a comparative analysis using the publicly available 7Scenes indoor dataset \cite{glocker2013real}, which features a diverse range of scenes, complex camera trajectories, and varied textures. To thoroughly assess the generalizability of GIS, we selected benchmark methods based on different underlying mechanisms, including the CNN-based PoseNet \cite{kendall2015posenet}, the LSTM-based LSTM PN \cite{walch2017image}, and the attention-based TransPoseNet \cite{shavit2024learning}. We first analyzed the training duration and pose estimation accuracy of each method, both before and after integrating GIS. Specifically, we measured the total training time (in minutes) and per-frame inference time (in milliseconds) for each method. Additionally, we compared the robustness of these benchmarks to extremely sparse training data, both with and without GIS. To ensure consistency, all models were trained on the same NVIDIA L20 server.

As shown in Table \ref{tab:Generalization}, the addition of the GIS improved the average translation and rotation accuracy across all models by $10.8\%$ and $4.99\%$, respectively, and increased the average training speed by $87.36\%$. Specifically, PoseNet \cite{kendall2015posenet} saw improvements of $15.9\%$ in translation and $13.6\%$ in rotation accuracy, with training time reduced by $88.32\%$. LSTM PN \cite{walch2017image} demonstrated $6.5\%$ improvements in translation and $9.3\%$ in rotation accuracy, also with training time reduced by $70\%$. For the state-of-the-art model TransPoseNet \cite{shavit2024learning}, the GIS  demonstrated $5.56\%$ improvements in translation and reduced the training time by $92.13\%$, though rotation error increased by 0.93°. This discrepancy may stem from the fact that, as noted by the authors of TransPoseNet \cite{shavit2024learning}, their results involve fine-tuning the orientation head with a latent position before the position Transformer-Encoder. In contrast, our approach, designed with time efficiency and practical deployment in mind, achieves results through a single-phase training process without additional fine-tuning.


Additionally, we compared the ability of the Global Information Selector (GIS) to enhance the robustness of various Absolute Pose Regression (APR) methods on extremely sparse training sets. As shown in Table \ref{tab:NLMB_sparsity}, we evaluated pose estimation errors on the entire test set after uniformly sampling 100\%, 10\%, and 5\% of the training set. The arrow direction indicates the increase/decrease in pose error of Benchmark-GIS relative to Benchmark under the same sparsity conditions. The experimental results demonstrate that the use of GIS significantly mitigates the degradation in translation error under all sparsity conditions, and in most cases, it also alleviates the degradation in rotation error. All benchmarks with GIS show reduced average translation and rotation errors across all experiments.

Finally, we assessed the computational complexity introduced by GIS. When the sequence output from the final feature extractor of a model has a batch size \( B \) with \( D \) channels, GIS adds a computational complexity of only \( O(BLDN) \), where \( L \) is set to 2 and the state dimension \( N \) is set to 16. This complexity scales linearly with the number of channels in the final feature extractor across different benchmarks. Table \ref{tab:model_size} provides a summary of the model sizes before and after incorporating GIS across various benchmarks, demonstrating that the introduction of GIS leads to an average model size increase of 38.44 MB. Notably, for the state-of-the-art TransPoseNet \cite{shavit2024learning}, this increase was limited to just 1.67 MB. These results underscore GIS as a lightweight and adaptable module that significantly improves model performance and convergence speed. Although incorporating GIS results in a slight increase in model size, which leads to marginally longer inference times compared to the benchmark (ranging from 0 ms per frame for PoseNet to a maximum of 1.16 ms per frame for TransPoseNet \cite{shavit2024learning}), the overall benefits substantially outweigh these minor costs. In summary, unlike attention mechanisms that incur quadratic growth in computational complexity, GIS enhances model accuracy, reduces training time, and improves robustness to sparse training data, all with only a linear increase in complexity \( O(BLDN) \). This highlights the efficiency and practicality of GIS in advancing end-to-end localization models.

\begin{figure*}[h]
    \centering
    \caption{ Principal Component Analysis (PCA) to visualize the feature extraction performance by plotting the translation and rotation feature maps of TransPoseNet \cite{shavit2024learning} with and without GIS on the 7Scenes Dataset \cite{glocker2013real}.}
    \includegraphics[width=\textwidth]{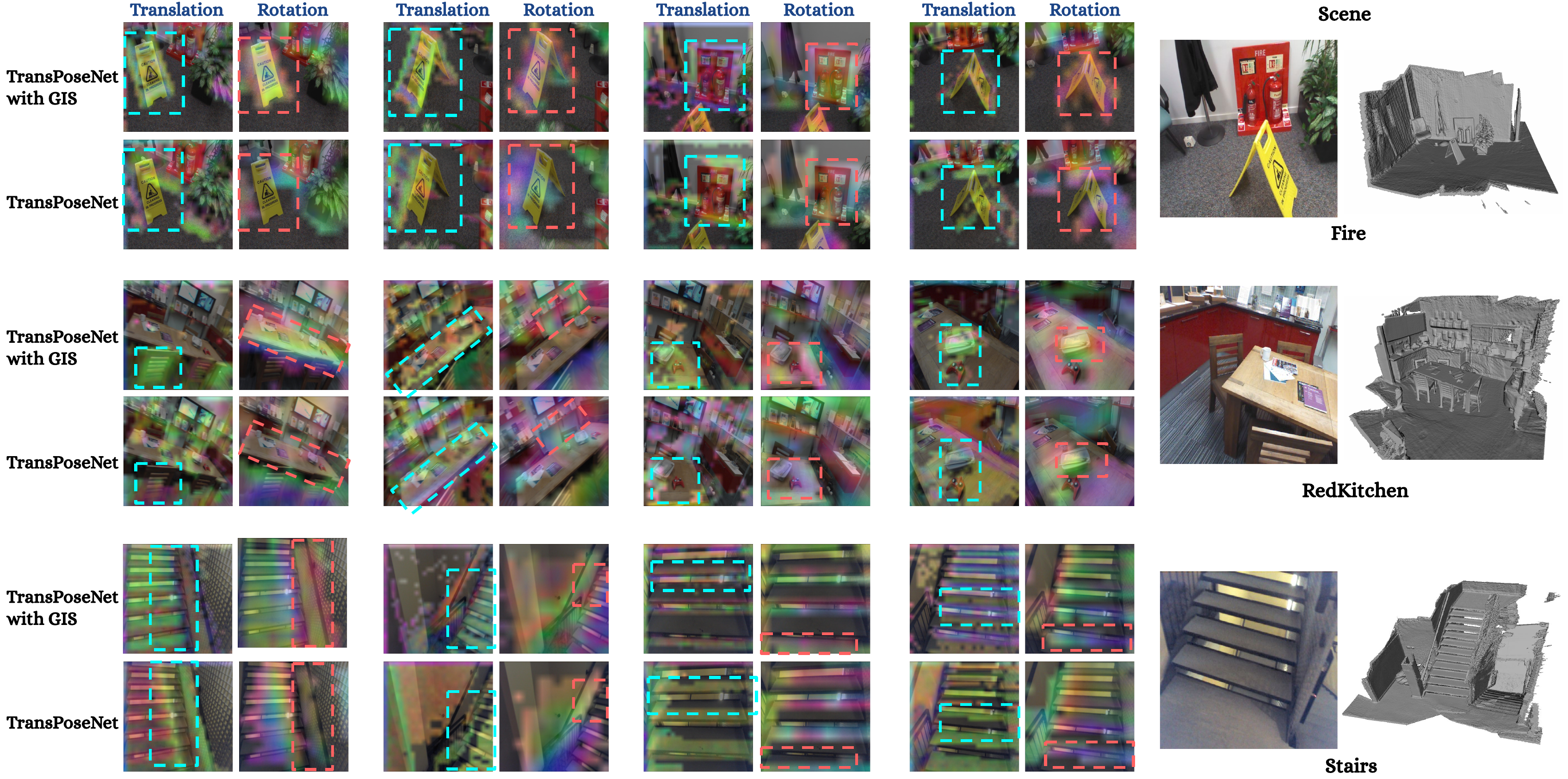}
    \label{fig:PCA Feature Maps}
\end{figure*}

\textbf{Qualitative Evaluation:} To enhance interpretability, we employed Principal Component Analysis (PCA) to visualize the feature extraction performance of different models by plotting the feature maps generated by the output layer of the position encoders. We used the state-of-the-art TransPoseNet (2024) \cite{shavit2024learning} as a benchmark, comparing its final position encoder layer's PCA visualizations with those obtained after incorporating our Global Information Selectors (GIS) into MambaLoc. As shown in Figure \ref{fig:PCA Feature Maps}, we analyzed the feature maps from the position encoders for query images across various scenes in the 7Scenes dataset \cite{glocker2013real}. Both methods effectively extract features for position and orientation tasks; however, the features extracted by MambaLoc more precisely capture and align with object edges, resulting in clearer and more focused feature maps. In contrast, the PCA visualizations for TransPoseNet reveal that its attention is more dispersed, with less accurate target locking and the introduction of noise and blurred boundaries. This suggests that MambaLoc offers enhanced interpretability by more efficiently isolating and emphasizing critical features, leading to better performance in feature extraction tasks.

 \begin{figure}[h]
    \centering
\includegraphics[width=0.5\textwidth]{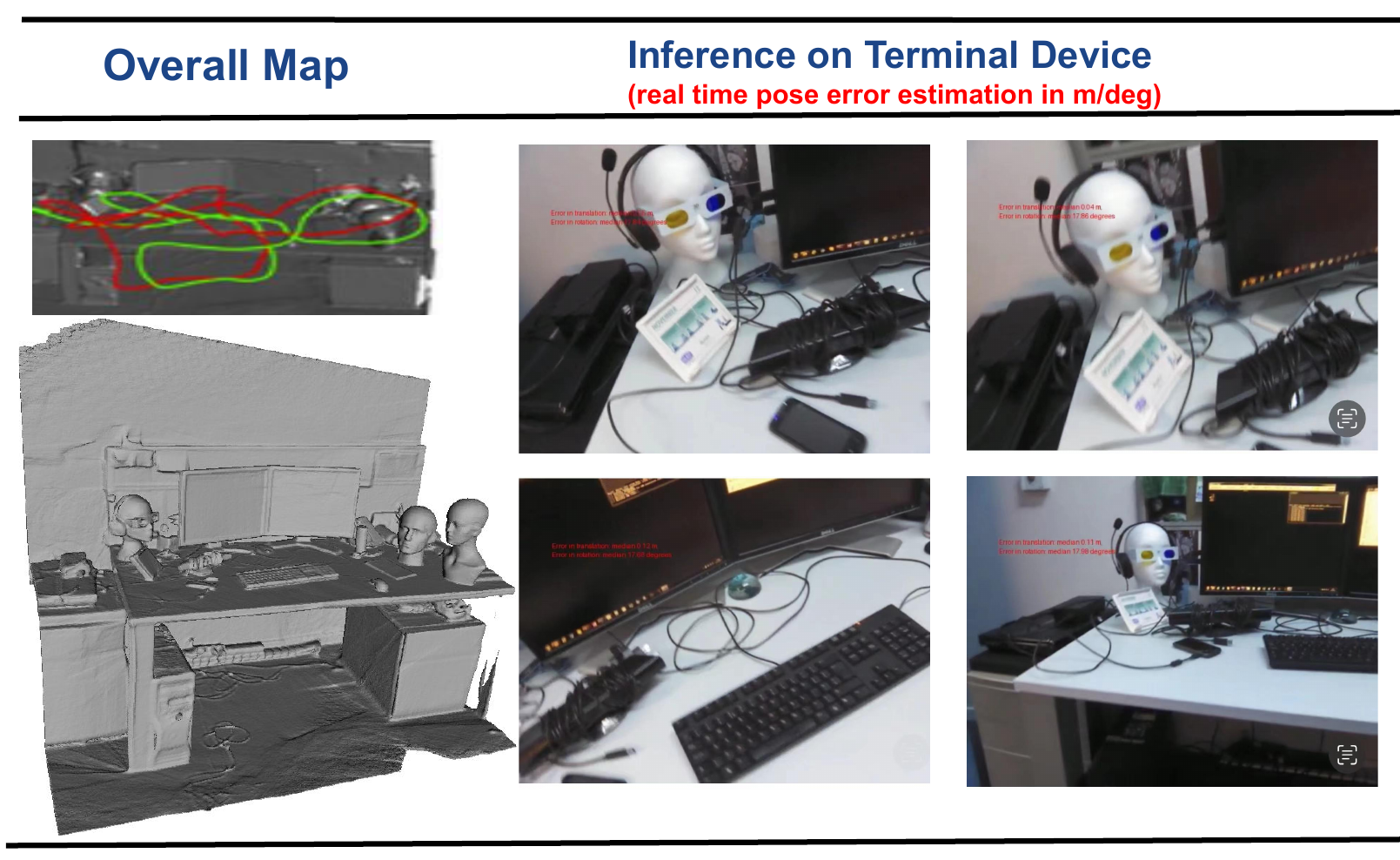}
   
    \caption{Real-Time Inference Demonstration of MambaLoc Deployed on PC}
    \label{fig:Terminal_demo}
\end{figure}

\subsection{Comparative Analysis of MambaLoc and Other Visual Localization Models for Robustness in Sparse Training Environments}

In the field of visual localization, especially in the context of edge-cloud collaborative localization and sparse keyframe SLAM loop closure detection, the ability of localization models to effectively handle sparse training data is crucial. Traditional camera localization networks and bag-of-words models used in SLAM systems often struggle with robustness when confronted with sparse viewpoints. MambaLoc addresses this issue by leveraging the inherent parameter sparsity of the SSM model, coupled with advanced feature compression techniques, to ensure robustness in data-limited environments. This capability enables MambaLoc to maintain high performance even in scenarios characterized by significant data sparsity.

To evaluate the robustness of MambaLoc in such scenarios, we conducted a comparative assessment, benchmarking it against methods based on different mechanisms, including the latest attention-based TransPoseNet \cite{shavit2024learning} and the classic CNN-based PoseNet \cite{kendall2015posenet}. Each model was trained on the 7Scenes dataset \cite{glocker2013real} with training sets reduced to 1/2, 1/3, and 1/10 of the full dataset, and their performance was evaluated on the complete test set. Additionally, we investigated MambaLoc's operational limits by training it on a highly sparse dataset (1/20 of the full training set). These experimental setups are practically significant for scenarios where rapid initialization of a preliminary map with minimal imagery is essential, particularly in unfamiliar environments requiring quick adaptation.

As shown in Table \ref{tab:sparsity_performance_comparison}, in most scenarios, all models exhibit a decline in performance as the training set becomes increasingly sparse. However, it is noteworthy that even with only 5\% of the training set, MambaLoc still achieved the smallest translation and rotation errors. Furthermore, during our experiments, we observed that MambaLoc was able to generate an initial environmental map for the heads scene in the 7Scenes dataset, which contains only 1,000 images, within 22.8 seconds using just 0.05\% of the complete dataset, even in extremely sparse scenarios.

The experiments in this section demonstrate MambaLoc's exceptional robustness under sparse data conditions, highlighting its suitability for real-world applications that require efficient localization with limited visual inputs. Additionally, we explored the limits of MambaLoc's performance in extremely sparse scenarios and demonstrated the model's ability to achieve rapid localization with minimal data input, which is particularly valuable for the quick construction of preliminary environmental maps.

\begin{figure*}[h]
    \centering

   \includegraphics[width=\textwidth]{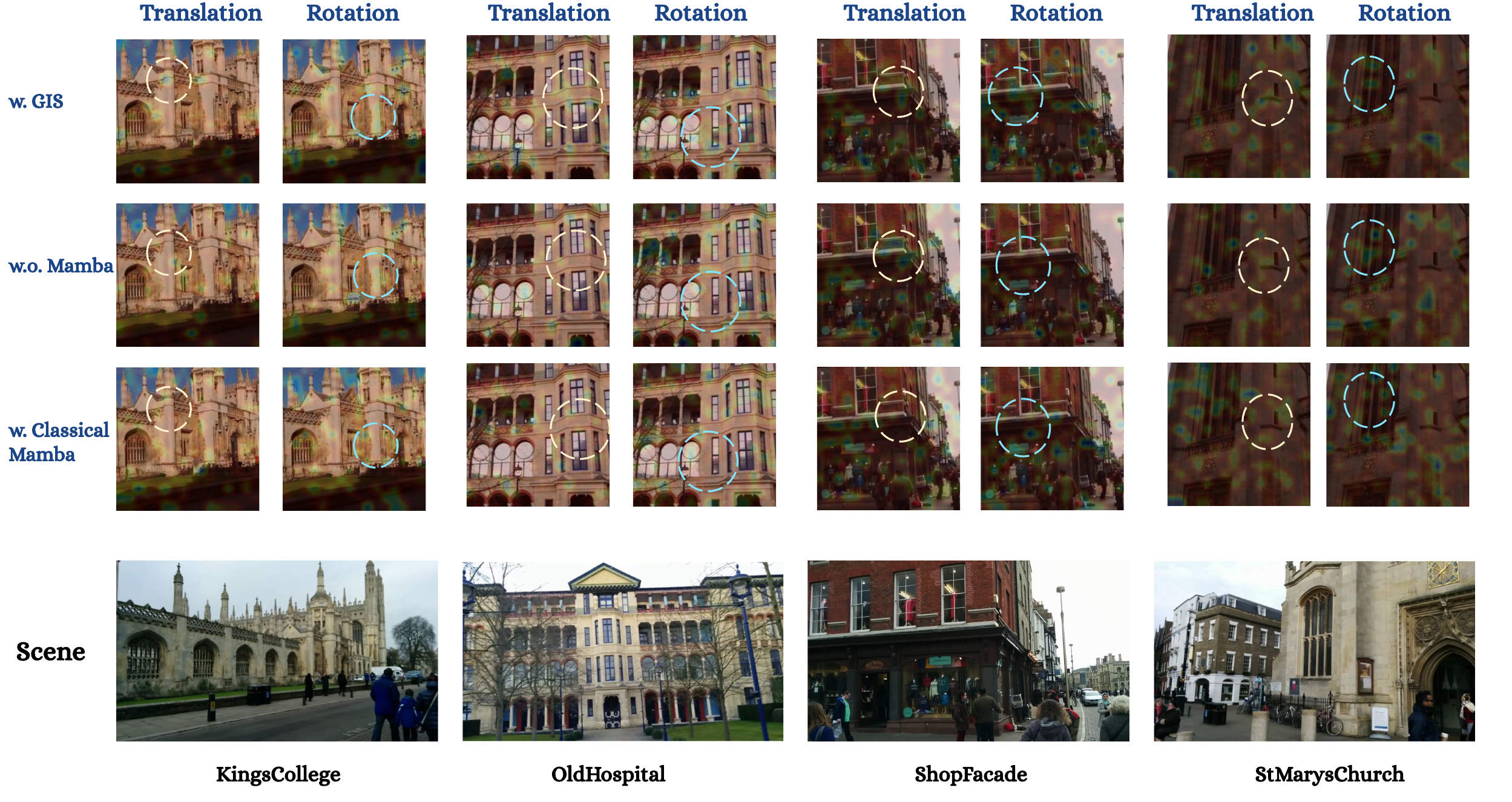}
 \caption{\textbf{Visualization of Global Feature Extraction with Global Information Selector (GIS)}: This figure uses CAM attention heatmaps to compare the global feature extraction capabilities of three MambaLoc configurations: with the  Global Information Selector (MambaLoc w. GIS), with the Classical Mamba Block (MambaLoc w. Classical Mamba), and without any Mamba Block (MambaLoc w.o. Mamba). While all configurations extract features for tasks such as position (via corner-like features) and orientation (via edge-like features), MambaLoc (top row) demonstrates more accurate and clearer feature extraction. In contrast, the attention heatmaps for MambaLoc w. Classical Mamba (middle) and MambaLoc w.o. Mamba (bottom) exhibit more noise.
}
\label{fig:CAM_heatmap}
\end{figure*}

\subsection{Performance Evaluation of MambaLoc Deployment on Terminal Devices}

To address the challenge of deploying networks with typically large parameter sizes in hardware-constrained mobile environments while maintaining accuracy, we employed the hybrid knowledge distillation method described in Section \ref{MambaLoc Application on Terminal Devices}. This method integrates feature-based and logits-based techniques to deploy MambaLoc on terminal devices. By doing so, we circumvent the limitations associated with edge device configurations and network conditions, enabling rapid local image processing. Table \ref{tab:Distillation_performance} presents a comparison of pose accuracy before and after distilling MambaLoc onto our modified student network, based on EfficientNet-B0 \cite{tan2019efficientnet} (network design detailed in Section \ref{MambaLoc Application on Terminal Devices}), using this hybrid knowledge distillation approach on the 7Scenes dataset. The experimental results demonstrate that our method not only reduced the model size by 47.62\% but also decreased the translation and rotation errors by 88.24\% and 93.80\%, respectively. We then compiled the distilled model using ONNX Runtime, ensuring compatibility across any terminal device that supports ONNX Runtime, such as iOS and Android mobile phones, as well as PCs. For a detailed demonstration of MambaLoc's deployment on terminal devices and real-time 6-DoF pose inference, please refer to Figure \ref{fig:Terminal_demo} and the accompanying video at https://youtu.be/sAvxB7bfR2s.

\begin{table*}[t]

    \centering
    \caption{\textbf{Performance of MambaLoc after Distillation on Terminal Testing: Pose Accuracy in Terms of Median Translation Error (m), Rotation Error (°) and Model Size (Mb)}}
    \label{tab:Distillation_performance}
    \tabcolsep 5pt    
    \begin{tabular}{@{}c|ccccccc|c|ccc@{}}
        \toprule
        \begin{tabular}[c]{@{}c@{}}MambaLoc State /Scenes\end{tabular} &
        Heads &
        Fire &
        Kitchen &
        Chess &
        Office &
        Pumpkin &
        Stairs &
        \begin{tabular}[c]{@{}c@{}}Avg. Pose \\Error\end{tabular} &
        \begin{tabular}[c]{@{}c@{}}Model Size \\ (Mb)\end{tabular} \\ \midrule
        \textbf{Before Distillation} &
			\begin{tabular}[c]{@{}c@{}}0.13/12.30 \end{tabular} &
			\begin{tabular}[c]{@{}c@{}}0.25/10.79\end{tabular} &
			\begin{tabular}[c]{@{}c@{}}0.17/7.85\end{tabular} &
			\begin{tabular}[c]{@{}c@{}}0.09/6.14\end{tabular} &
			\begin{tabular}[c]{@{}c@{}}0.17/6.58\end{tabular} &
			\begin{tabular}[c]{@{}c@{}}0.16/6.56\end{tabular} &
			\begin{tabular}[c]{@{}c@{}}0.23/10.74\end{tabular} &
			\begin{tabular}[c]{@{}c@{}}0.17/8.71\end{tabular} &
            \begin{tabular}[c]{@{}c@{}} 42.24 \end{tabular} \\
        \textbf{After Distillation} &
            \begin{tabular}[c]{@{}c@{}} \textbf{0.01/0.44}\end{tabular} &
            \begin{tabular}[c]{@{}c@{}} \textbf{0.02/0.46}\end{tabular} &
            \begin{tabular}[c]{@{}c@{}} \textbf{0.03/0.71}\end{tabular} &
            \begin{tabular}[c]{@{}c@{}} \textbf{0.02/0.51}\end{tabular} &
            \begin{tabular}[c]{@{}c@{}} \textbf{0.02/0.63}\end{tabular} &
            \begin{tabular}[c]{@{}c@{}} \textbf{0.02/0.59}\end{tabular} &
            \begin{tabular}[c]{@{}c@{}} \textbf{0.02/0.41}\end{tabular} &
            \begin{tabular}[c]{@{}c@{}} \textbf{0.02/0.54}\end{tabular} &
            \begin{tabular}[c]{@{}c@{}}\textbf{22.69} \end{tabular} &
            \begin{tabular}[c]{@{}c@{}} \end{tabular} \\ \bottomrule
    \end{tabular}
\end{table*}

\begin{table*}[t]

    \centering
    \caption{\textbf{Comparative Ablation Study on the Global Information Selector (GIS)}: Evaluation of pose estimation accuracy and training duration across three MambaLoc configurations for 7-Scenes \cite{glocker2013real} Dataset. The analysis presents Pose Regression Results, Total Training Time, and Average Inference Time in Terms of Median Translation Error (m), Rotation Error (°), Training Time (minutes), and Inference Time per Frame (minutes), all averaged for each dataset.}
    \label{tab:Comparative_Ablation_Study}
    \tabcolsep 5pt    
    \begin{tabular}{@{}c|cccccccccc@{}}
        \toprule
        \begin{tabular}[c]{@{}c@{}}Method/Scenes\\ (\#trainset)\\ (\#testset)\end{tabular} &
        \begin{tabular}[c]{@{}c@{}}Heads\\ \#1000\\ \#1000\end{tabular} &
        \begin{tabular}[c]{@{}c@{}}Fire\\ \#2000\\ \#2000\end{tabular} &
        \begin{tabular}[c]{@{}c@{}}Kitchen\\ \#7000\\ \#1000\end{tabular} &
        \begin{tabular}[c]{@{}c@{}}Chess\\ \#4000\\ \#2000\end{tabular} &
        \begin{tabular}[c]{@{}c@{}}Office\\ \#6000\\ \#4000\end{tabular} &
        \begin{tabular}[c]{@{}c@{}}Pumpkin\\ \#4000\\ \#2000\end{tabular} &
        \begin{tabular}[c]{@{}c@{}}Stairs\\ \#2000\\ \#1000\end{tabular} &
        \begin{tabular}[c]{@{}c@{}}Avg. \\Pose Error \\ (m/°)\end{tabular} &
        \begin{tabular}[c]{@{}c@{}}Avg.\\ Training \\ Time (min)\end{tabular} &
        \begin{tabular}[c]{@{}c@{}}Avg.\\ Inference \\ Time (mim/f)\end{tabular} \\ \midrule
        \textbf{w.o. Mamba} &
			0.13/12.70 &
			\textbf{0.24}/\textbf{10.60} &
			0.19/\textbf{6.75} &
			\textbf{0.08}/\textbf{5.68} &
			\textbf{0.17}/\textbf{6.34} &
			0.17/\textbf{5.60} &
			0.30/\textbf{7.02 }&
			0.18/\textbf{7.78}&
			178 & \textbf{0.167e-3} \\
        \textbf{w. Classical Mamba} &
        \textbf{0.12}/12.84&
        0.25/10.63 &
        0.18/7.19 &
        0.09/6.55 &
        \textbf{0.17}/7.10&
       \textbf{ 0.16}/5.88 &
        \textbf{0.23}/9.68 &
       \textbf{ 0.17}/8.55 &
        27 &
         0.191e-3  \\
        \textbf{w. GIS (ours)} &
			\begin{tabular}[c]{@{}c@{}}0.13/\textbf{12.30} \end{tabular} &
			\begin{tabular}[c]{@{}c@{}}0.25/10.79\end{tabular} &
			\begin{tabular}[c]{@{}c@{}}\textbf{0.17}/7.85\end{tabular} &
			\begin{tabular}[c]{@{}c@{}}0.09/6.14\end{tabular} &
			\begin{tabular}[c]{@{}c@{}}\textbf{0.17}/6.58\end{tabular} &
			\begin{tabular}[c]{@{}c@{}}\textbf{0.16}/6.56\end{tabular} &
			\begin{tabular}[c]{@{}c@{}}\textbf{0.23}/10.74\end{tabular} &
			\begin{tabular}[c]{@{}c@{}}\textbf{0.17}/8.71\end{tabular} &
			\begin{tabular}[c]{@{}c@{}}\textbf{14}\end{tabular}
            & 0.197e-6 \\

        \bottomrule
    \end{tabular}
\end{table*}

\subsection{Ablation Study on the Effectiveness of Global Information Selector (GIS)}

\textbf{Quantitative Evaluation:} To demonstrate the effectiveness of the proposed Global Information Selector (GIS), we conducted an ablation study comparing pose estimation accuracy, total training time, and average inference time per frame across three models: MambaLoc with the Global Information Selector (MambaLoc w. GIS), MambaLoc with the Classical Mamba Block (MambaLoc w. Classical Mamba), and MambaLoc without any Mamba or GIS (MambaLoc w.o. Mamba). This evaluation was performed using the 7-Scenes dataset \cite{glocker2013real}. As shown in Table \ref{tab:Comparative_Ablation_Study}, the average pose estimation errors and inference speeds among the three models are comparable. However, compared to the complete MambaLoc with the Global Information Selector, the model using the Classical Mamba without non-local mechanisms exhibited a 84.83\% increase in training time cost, and the model without any Mamba module showed a 92.13\% increase in training time cost. These experimental results demonstrate that the GIS module significantly accelerates training speed without a substantial compromise in pose estimation accuracy.

\textbf{Qualitative Evaluation:} The interpretability of self-supervised deep learning models for autonomous vehicles (AVs) is crucial for validating that predictions are grounded in intuitive geometric understanding of depth and self-motion. It also plays a significant role in increasing user trust and ensuring accountability to regulatory bodies. To assess this interpretability, we employed heatmaps to visualize the attention weights across different models. As illustrated in Figure \ref{fig:CAM_heatmap}, the position and orientation encoders of all three models are capable of focusing on distinct image features pertinent to their specific tasks. For instance, the network emphasizes corners and distinctive points to estimate relative translation, while it targets edges and lines for relative rotation estimation. Notably, the integration of MambaLoc with the Global Information Selector (w. GIS) exhibited superior precision in feature targeting, thereby substantially enhancing both the model’s performance and interpretability.

\section{Conclusion}
In this article, we addressed the critical challenge of achieving reliable positioning in the context of terminal devices and edge-cloud IoT systems, such as autonomous vehicles and augmented reality, where significant training costs and the necessity of densely collected data often hinder progress. To tackle these issues, we introduced MambaLoc, an innovative model that applies the Selective State Space (SSM) model to visual localization. MambaLoc achieves high training efficiency by leveraging the SSM model’s capabilities, which include efficient feature extraction, rapid computation, and optimized memory usage. Additionally, the sparsity of the model’s parameters reduce computational overhead and enhance robustness, enabling MambaLoc to maintain high accuracy and perform reliably even with limited training data. Furthermore, we introduce the Global Information Selector (GIS), which leverages Mamba's capabilities by employing a set of learnable parameters to selectively compress input features into a more compact and effective global feature representation. Additionally, we integrate Mamba's Hardware-aware Algorithm to enhance training efficiency. This approach allows a single Mamba layer to function as a Non-local Neural Network, effectively capturing long-range dependencies with minimal layers and accelerating convergence. 

Our comprehensive experimental validation on public indoor and outdoor datasets first highlights the effectiveness of our model and further demonstrates the versatility of GIS when integrated with various existing localization models. For example, in our experiments with the Heads Scene of the 7Scenes dataset, MambaLoc used only 0.05\% of the training data and achieved accuracy comparable to leading methods in just 22.8 seconds. Additionally, MambaLoc’s deployment on end-user devices showcased its practical applicability, reinforcing its potential as a robust and efficient solution for visual localization. In summary, MambaLoc offers a significant advancement in visual localization, enhancing location services for terminal devices and edge-cloud IoT systems by providing a more efficient and reliable solution that overcomes the limitations of traditional methods.

\vfill

\end{document}